\documentclass[lettersize,journal]{IEEEtran}
\usepackage{amsmath,amsfonts}
\usepackage{algorithmic}
\usepackage{algorithm}
\usepackage{array}
\usepackage[caption=false,font=normalsize,labelfont=sf,textfont=sf]{subfig}
\usepackage{textcomp}
\usepackage{stfloats}
\usepackage{url}
\usepackage{verbatim}
\usepackage{graphicx}
\usepackage{cite}
\usepackage{siunitx}
\usepackage{mhchem}
\usepackage{diagbox}
\usepackage{multirow}
\usepackage{booktabs}
\usepackage{makecell}
\usepackage[export]{adjustbox}
\usepackage{arydshln}
\usepackage{color}
\usepackage[colorlinks,
            linkcolor=red,
            anchorcolor=blue,
            citecolor=green
            ]{hyperref}

\usepackage{stackengine}
\newcommand\xrowht[2][0]{\addstackgap[.5\dimexpr#2\relax]{\vphantom{#1}}}

\begin{document}

\title{Diverse and Lifespan Facial Age Transformation Synthesis with Identity Variation Rationality Metric}

\author{Jiu-Cheng Xie,~\IEEEmembership{member,~IEEE},
Jun Yang, Wenqing Wang, Feng Xu, Jian Xiong,~\IEEEmembership{member,~IEEE}, and Hao Gao, ~\IEEEmembership{member,~IEEE}

\thanks{This work was supported in part by National Nature Science Foundation of China under Grant 62301278, Grant 62371254, Grant 61931012; in part by Nature Science Foundation of Jiangsu Province of China under Grant BK20230362 and Grant BK20210594; and in part by Natural Science Research Start-up Foundation of Recruiting Talents of Nanjing University of Posts and Telecommunications under Grant NY222019 and Grant NY221019. \textit{(Corresponding author: Hao Gao.)}}
        
\thanks{Jiu-Cheng Xie, Jun Yang, and Hao Gao are with the School of Automation and School of Artificial Intelligence, Jian Xiong is with the School of Communications and Information Engineering, Nanjing University of Posts and Telecommunications, Nanjing 210003, China (e-mail: jiuchengxie@gmail.com; perryj200009@gmail.com; tsgaohao@gmail.com; jxiong@njupt.edu.cn).}

\thanks{Wenqing Wang is with the Centre for Vision, Speech and Signal Processing (CVSSP), University of Surrey, GU2 7XH Guildford, U.K. (e-mail: wenqing.wang@surrey.ac.uk).}

\thanks{Feng Xu is with the School of Software and BNRist, Tsinghua University, Beijing 100190, China (email: xufeng2003@gmail.com).}
}

\markboth{Journal of \LaTeX\ Class Files,~Vol.~14, No.~8, August~2021}%
{Shell \MakeLowercase{\textit{et al.}}: A Sample Article Using IEEEtran.cls for IEEE Journals}


\maketitle

\begin{abstract}
Face aging has received continuous research attention over the past two decades. Although previous works on this topic have achieved impressive success, two longstanding problems remain unsettled: 1) generating diverse and plausible facial aging patterns at the target age stage; 2) measuring the rationality of identity variation between the original portrait and its syntheses with age progression or regression. In this paper, we introduce $\rm{\textbf{DLAT}}^{\boldsymbol{+}}$ to realize Diverse and Lifespan Age Transformation on human faces, where the diversity jointly manifests in the transformation of facial textures and shapes. Apart from the diversity mechanism embedded in the model, multiple consistency restrictions are leveraged to keep it away from counterfactual aging syntheses. Moreover, we propose a new metric to assess the rationality of Identity Deviation under Age Gaps (IDAG) between the input face and its series of age-transformed generations, which is based on statistical laws summarized from plenty of genuine face-aging data. Extensive experimental results demonstrate the uniqueness and effectiveness of our method in synthesizing diverse and perceptually reasonable faces across the whole lifetime. 

\end{abstract}


\section{Introduction}
\IEEEPARstart{W}{ith} a human portrait, facial age transformation synthesis aims to simulate the corresponding appearance at the target age stage, which is user-specified. In the past two decades, this topic has attracted numerous attention in the computer vision community. The reasons for continuous concerns partially come from its relevant applications: special effects production in film animation entertainment and searching for people lost for many years. Advances in these fields will bring substantial economic and social benefits.
\par

Most early efforts at facial age transformation study the transition between adults and the elderly, during which the facial changes are mainly at the texture level (e.g., the increase or decrease of the wrinkles, the changes in hair color and amount). Benefiting from the rapid progress in generative networks (e.g., from StyleGAN1 to StyleGAN3 \cite{karras2019style,karras2020analyzing,karras2021alias}) and the open access of face datasets (e.g., FFHQ \cite{Database:FFHQ}) characterized by comprehensive age coverage from infancy to senectitude along with a large number of high-resolution portraits under each certain period, LATS \cite{or2020lifespan} comes out as the groundbreaking work in realizing lifelong age transformation synthesis. After that, several researchers proposed respective methods \cite{he2021disentangled,hou2022lifelong,alaluf2021only,hsu2022agetransgan,gomez2022custom,10458950}, pushing research in this field to new boundaries. 

While these algorithms achieve impressive performance, they share a common limitation with previous works: \textit{the generated faces of the given subject under the target age condition look the same}. In other words, these models can just simulate deterministic age regression or progression effects, which is contrary to the universal facial aging rules in reality. In short, human growth is affected by internal factors dependent on age-related genes and many external factors, including lifestyle, climate, professions, etc. Uncertainty in these aspects determines that there should be infinite candidates for a subject's possible appearance at the specified age stage. Liu et al. \cite{liu2022multimodal} and Li et al. \cite{li2023pluralistic} made pioneering attempts to address the diverse aging problem. While they obtain impressive aging effects, a common characteristic of them is the dependence of another facial image to provide age-relevant information, which is the full \cite{liu2022multimodal} or partial \cite{li2023pluralistic} origin of diverse signals. The design based on a real reference face may hinder their wide real-world applications.


\par

Most of the time, we do not have the input subject's portrait at the target age phases in the real world, so it is difficult to accurately evaluate the reasonability of identity variation between the source and synthesized faces. To achieve quantitative evaluation, some works \cite{li2021continuous,hsu2022agetransgan,liu2019attribute,yang2019learning} take an alternative scheme through conducting verification tests between input and generated faces, where the similarity score above a pre-defined threshold will be regarded as a successful verification and the opposite as a failed case. Regarding performance evaluation, \textit{they simply equate the higher verification rate with more reasonable identity preservation between original and age-transformed face pairs}. We argue that this equivalence does not hold. In alignment with our opinion, \cite{liu20213} also reported the following observation according to their experiments: ``As the time interval of age progression increases, both conﬁdence scores and veriﬁcation rates gradually decrease.”  
\par

This paper presents $\rm{DLAT}^{\boldsymbol{+}}$ for \textbf{D}iverse and \textbf{L}ifespan \textbf{A}ge \textbf{T}ransformation to address the above-mentioned issues. The full model is composed of two sub-networks, namely $\rm{DLAT}_{img}$ and $\rm{DLAT}_{lmk}$. The former can realize the photorealistic simulation of the given subject at desired age stages across the whole life cycle, but with diverse aging patterns mainly manifesting at the texture level. The latter is used to compensate for the shortage of facial shape diversity, where this rectification is achieved by leveraging the transformation parameters yielded by $\rm{DLAT}_{lmk}$ to warp the output face obtained from $\rm{DLAT}_{img}$. A shared principle of diversity mechanism is embedded in these two networks during their respective model's training processes. In short, this mechanism first constructs the mapping from noise to age latent codes associated with specific age stages. Then, it forces the generator to synthesize faces located in expected age phases via conditioning on corresponding age codes. After this procedure, the connection between noise and multiple age-related latent distributions has been established. Then, various aging effects can be achieved through random noise sampling. Noting that, we devise a series of constraints based on identity, race, and pose consistency between corresponding face pairs before and after age transition to protect the model from excessive and unreasonable facial aging diversity. In addition, we analyze the variations of identity similarity among portraits at different age stages from the same subjects, which are contained in four public real-life datasets with accurate age labels. As a result, we obtain the general laws of identity similarity changes across ages in reality. Based on this rule, we design an \textbf{I}dentity \textbf{D}eviation under \textbf{A}ge \textbf{G}aps (IDAG) metric, which assesses identity variation rationality during facial age transformation. 
\par

To summarize, the contributions of this paper are as follows:
\begin{itemize}
    \item A new method for synthesizing diverse and plausible facial aging effects, where the target age stages cover the whole life cycle. Concretely, the proposed diversity mechanism is separately deployed to the generations of facial texture and shape, and then respective synthetic intermediates are fused through image warping. Besides, consistency restrictions of race, pose, and identity are applied to the model, making aging diversity within an appropriate range.
    \item A new metric for measuring the identity variation rationality between source faces and corresponding age-transformed generations, which essentially compares the identity change trends synthesized with universal variation rules in the real world.
    \item Extensive experiments demonstrate our method's unique ability to realize diverse age transformations compared to the latest techniques. In addition, under more challenging settings where gaps between source and target age stages are decades, our approach shows better identity shit reasonability and lower age transition errors than others.
\end{itemize}

\section{Related Work}
\subsection{Facial Age Transformation}
Early efforts at facial age transformation predominantly relied on traditional computer vision methods. These approaches often employ techniques such as geometric morphing \cite{todd1980perception,golovinskiy2006statistical,tazoe2012facial}, texture synthesis \cite{suo2009compositional,wu1994plastic,bando2002simple}, and statistical models to simulate age progression or regression \cite{tiddeman2001prototyping,kemelmacher2014illumination,shu2015personalized,wang2016recurrent}. However, these methods were limited in capturing intricate facial details and often produced unrealistic results. The advent of deep generative models (DGMs) marked a significant paradigm shift in facial age transformation research. Notable contributions include the works of Zhang et al. \cite{zhang2017age} and Antipov et al. \cite{antipov2017face}, where conditional adversarial autoencoder and conditional generative adversarial networks (cGANs) were employed respectively to generate facial images conditioned on age labels. These two pioneering works demonstrated the capacity of DGMs to produce high-quality, age-transformed faces with improved realism. Subsequent researchers have explored various modifications and extensions of conditional generative architectures to enhance their performance in facial age transformation tasks \cite{liu2017face,wang2018face,yang2018learning,he2019s2gan,liu20213,Chen2023faceaging}. 

Recent research cares about more challenging yet practical settings, such as 3D face aging \cite{Yiqiang2022TCSVT}, high-resolution aging results \cite{yao2021high,despois2020agingmapgan,zoss2022production}, lifelong aging synthesis \cite{or2020lifespan,alaluf2021only,hou2022lifelong,he2021disentangled,hsu2022agetransgan}, background structure \cite{gomez2022custom}, temporally consistent re-aging in videos \cite{muqeet2023video}, pluralistic aging \cite{liu2022multimodal,li2023pluralistic} and personality preservation \cite{li2021continuous,makhmudkhujaev2021re,10458950}. Compared to those prior works, our paper is distinct from them in three aspects: (1) introducing an algorithm for diverse and life-long age transformation without depending on real reference faces; (2) proposing the first metric for assessing the rationality of identity deviation on age-shifted synthetic faces; (3) probing into the beneficial impacts of diverse and lifespan age transformation on downstream applications conceptually and experimentally.

\subsection{Controllable Face Editing}
Face aging is closely connected to controllable face editing, which aims to alter specific attributes of a facial image (such as gender, hairstyle, expression, etc.) while keeping others unchanged. This task has been extensively studied in the past few years \cite{he2019attgan,peng2021unified,li2022mat,xu2022transeditor,li2021swapinpaint,wang2023clip2gan,liu2022deepfacevideoediting,wei2023hairclipv2,pernuvs2023maskfacegan,shen2020interfacegan,sun2022ide,zhu2023linkgan,liu2022towards}. For a more detailed survey in this field, please refer to \cite{liu2023gan}. According to their different operating modes, we roughly classify those approaches into two categories: training models from scratch with given control conditions \cite{he2019attgan,peng2021unified,li2022mat,xu2022transeditor,li2021swapinpaint} or manipulating relevant latent representations in a pre-trained GAN's feature space \cite{wei2023hairclipv2,wang2023clip2gan,liu2022deepfacevideoediting,shen2020interfacegan,sun2022ide,zhu2023linkgan,pernuvs2023maskfacegan,liu2022towards}. Under this taxonomy, our method belongs to the former. Although approaches of the latter escape from heavy training burdens, they need to afford the costs of rough attribute editing. For example, there are only binary choices in their application to facial age transformation: younger or older, whereas specialist algorithms can achieve more fine-grained manipulation up to specific age stages.
\par

\section{METHOD}
\begin{figure*}[t]
\centering
\includegraphics[width=0.95\textwidth]{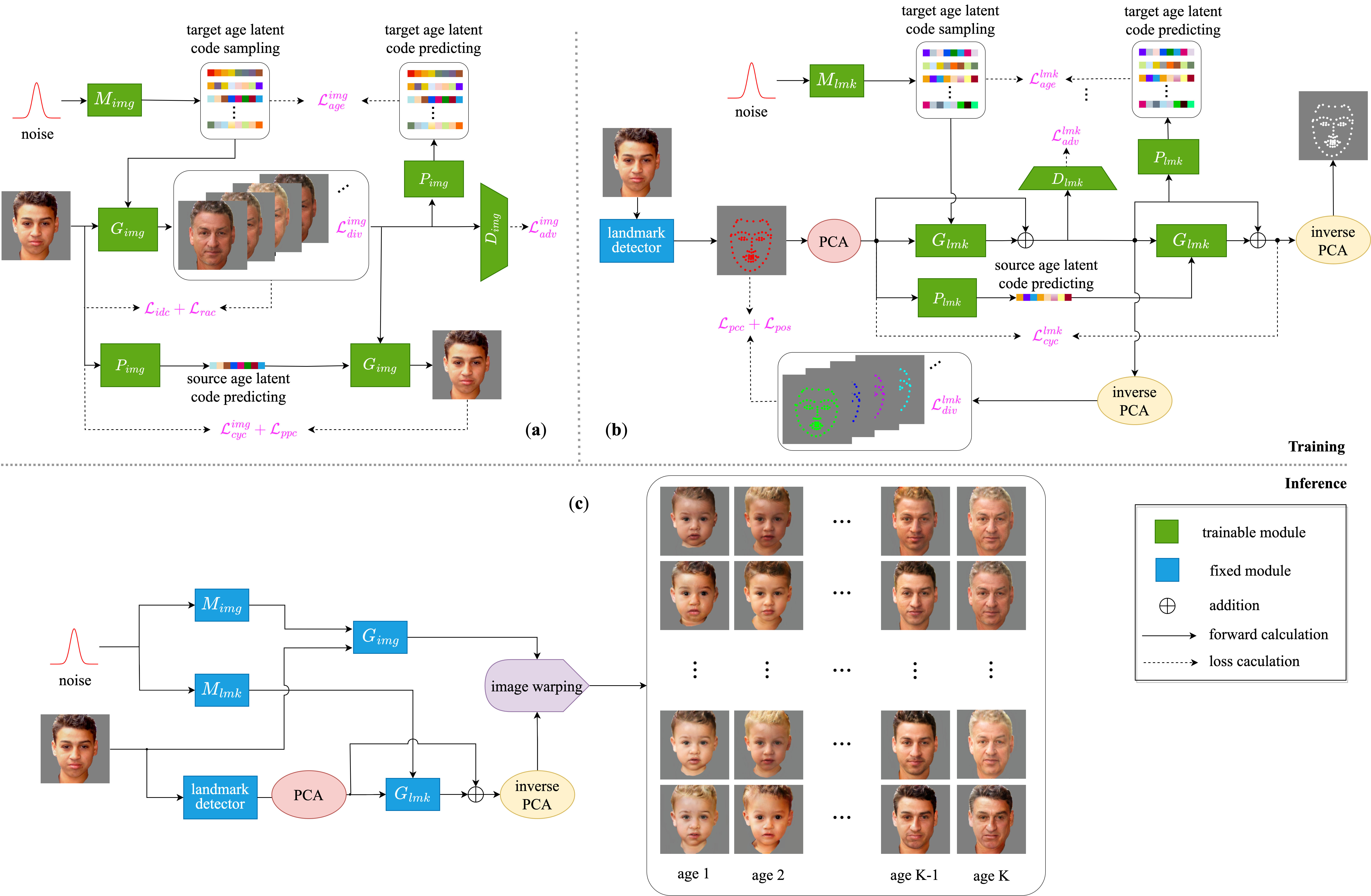}
\caption{An overview of our method. During the training stage, the proposed $\rm{DLAT}_{img}$ (a) and $\rm{DLAT}_{lmk}$ (b) are learned separately. While the former learns to generate multiple age-transformed faces under each specific age condition with diversity mainly manifesting in the facial texture, the latter purely focuses on synthesizing structural variety on faces represented by landmarks. Regarding the $\rm{DLAT}_{img}$ model, its age mapping module $M_{img}$ is used to learn the mapping from a set of randomly sampled noise to $K$ sets of age latent codes correlated with $K$ different age stages. Moreover, those codes under each specific age phase need to be distinctive from each other. This goal is achieved with the help of other three components. Concretely, a face generator $G_{img}$ uses the source portrait and conditions on age latent codes to generate new faces located in specified age stages, which is ensured by a face discriminator $D_{img}$ with an adversarial loss $\mathcal{L}_{adv}^{img}$.  Given these converted faces, a predictor $P_{img}$ is leveraged to estimate corresponding age latent codes, the conformance between which and the original ones is enforced by the $\mathcal{L}_{age}^{img}$. For those generated faces within the same age group, they are encouraged by $\mathcal{L}_{div}^{img}$ to be texturally different. To avoid pursuing unreasonable diversity, consistency of identity and race attributes are respectively required by $\mathcal{L}_{idc}$ and $\mathcal{L}_{rac}$ between generations and the source face. On the other hand, the age latent code predicted from the source image accompanied with age-shifted face syntheses are sent to the $G_{img}$ again with the supervision of $\mathcal{L}_{cyc}^{img}$ and $\mathcal{L}_{ppc}$, aiming at the reconstruction purpose. At the same time, similar diversity mechanisms and constraints designed for eliminating irrational transformation of facial geometry are imposed on the learning of $\rm{DLAT}_{lmk}$. After that procedure, pre-trained age mapping modules $M_{img}$ and $M_{lmk}$, as well as relevant generators $G_{img}$ and $G_{lmk}$, are picked out and integrated via image warping operations to form the $\rm{DLAT}^{\boldsymbol{+}}$ (c). In brief, it is an enhanced model capable of simulating age transformation effects across the life cycle on the given input faces with plausible variations both in facial appearance and shapes.}  
\label{pipeline}  
\end{figure*}
Given a facial photo of one person, our basic Diverse and Lifespan Age Transformation ($\rm{DLAT}_{img}$) model is able to synthesize \textit{diverse age progression or regression effects across the whole life cycle} on the subject's face. In addition, we introduce its counterpart based on facial landmarks ($\rm{DLAT}_{lmk}$) to focus on learning facial geometry deformation during the aging process. Collaborations between them through image-warping lead to a more powerful version dubbed $\rm{DLAT}^{\boldsymbol{+}}$, which can yield age-transformed faces with more variety and reasonability in terms of facial texture and shape. The complete pipeline of our method is given in Fig. \ref{pipeline}.
\par

\subsection{Diverse and Lifespan Age Transformation}
Denoting the given facial image as $\boldsymbol{X}$, our $\rm{DLAT}_{img}$ model takes it and random noise $\boldsymbol{n}$ as joint inputs and then produces a sequence of synthesized corresponding faces under user-predefined age stages $\{\boldsymbol{Y}_k\}_{k=1}^K$. Note that adjusting the input noise will lead to distinct but reasonable aging effects. This is formulated as
\begin{equation}
\{\boldsymbol{Y}_k\}_{k=1}^K=\rm{DLAT}_{img}(\boldsymbol{X}, \boldsymbol{n}), \qquad    \boldsymbol{n}\sim\mathcal{N}(\boldsymbol{\mu}, \boldsymbol{\Sigma})
\end{equation}
where $\boldsymbol{\mu}$ and $\boldsymbol{\Sigma}$ denote the mean and covariance matrix of the multivariate normal distribution chosen, and $K$ is the total number of different age phases considered. The training of $\rm{DLAT}_{img}$ involves four separate network modules, namely diverse age mapper $M_{img}$, face generator $G_{img}$, age latent code predictor $P_{img}$ and the image discriminator $D_{img}$. While in the inference stage, only the former two components are used. Fig. \ref{pipeline}(a) depicts the framework of the proposed $\rm{DLAT}_{img}$.
\par

\textit{Diverse age mapping module.} Our diverse age mapping module converts a piece of noise into a set of age latent codes, $\{\boldsymbol{u}_k\}_{k=1}^K=M_{img}(\boldsymbol{n})$, corresponding to non-overlapped and ordinal age stages. Through random noise sampling, this module is expected to produce various but correlated age latent codes belonging to the same age group. In practice, we employ an MLP structure for its implementation, where the number of outputs is set to $K$.
\par

\textit{Face generator module.} Our face generator uses two modalities of signals as its input: a face image (real/generated) and an age latent code (sampled/predicted). For example, given the source input face $\boldsymbol{X}$ and an age latent code $\boldsymbol{u}_k$, our generator can synthesize a possible target face $\boldsymbol{Y}_k=G_{img}(\boldsymbol{X}, \boldsymbol{u}_k)$, with desired age patterns. Here, we take an encoder-decoder form for the generator's implementation, which generally follows the previous work \cite{he2021disentangled}. The impact of age latent codes on extracted facial features is realized by "weight demodulation" \cite{karras2020analyzing}. The generator is encouraged to yield facial syntheses with distinct age transformation effects by conditioning on different age latent codes from the same age range. This goal is achieved by forcing the module to maximize: 
\begin{equation}
\label{div_loss}
\mathcal{L}_{div}^{img}=\mathbb{E}_{\boldsymbol{{X}},\boldsymbol{u}_k}[||G_{img}(\boldsymbol{X},\boldsymbol{u}_k^1)-G_{img}(\boldsymbol{X},\boldsymbol{u}_k^2)||],
\end{equation}
where $\boldsymbol{u}_k^1$ and $\boldsymbol{u}_k^2$ represent two different age latent codes sampled from the age stage $k$.
\par

\textit{Age latent code predictor module.} The predictor must accurately predict the corresponding age latent codes of real or generated facial images. We achieve this goal by forcing the predictor to give predictions of the age-transformed faces consistent with the sampled latent age codes used to synthesize them. This is formulated as
\begin{equation}
\begin{split}
\mathcal{L}_{age}^{img}=\mathbb{E}_{\boldsymbol{X},\boldsymbol{u}_k}[||P_{img}(G_{img}(\boldsymbol{X},\boldsymbol{u}_k))-\boldsymbol{u}_k||].   
\end{split}
\end{equation}
The effects of this module are two-fold: (a) making the face generator produce results without ignoring the input age information and (b) providing necessary age-related signals to realize image reconstruction constraints. The predictor's structure in our realization is the concatenation of multiple convolutional layers and $K$ parallel fully connected layers as outputs.
\par

\textit{Face discriminator module.} Similar to \cite{or2020lifespan,he2021disentangled}, here we employ a multi-task learning discriminator with $K$ output neurons. The $k$th branch makes binary classifications of the input face: whether the subject looks realistic and shows reasonable aging patterns at the $k$th age stage or not. The related adversarial learning objective is formulated as follows:
\begin{equation}
\begin{split}
\mathcal{L}_{adv}^{img}=&\mathbb{E}_{\boldsymbol{X}}[\log D_{img}(\boldsymbol{X})]+\\
&\mathbb{E}_{\boldsymbol{X}, \boldsymbol{u}_k}[\log(1- D_{img}(G_{img}(\boldsymbol{X},\boldsymbol{u}_k)))].
\end{split}
\end{equation}
\par

\subsection{Rectification of Facial Geometry Diversity}
Experiments show that diverse age transformation effects produced by the $\rm{DLAT}_{img}$ are mainly changes of facial texture while exhibiting few variations in facial shape. Thus, we devise a geometric version of Diverse and Lifespan Age Transformation model dubbed $\rm{DLAT}_{lmk}$, focusing on learning the deformation of facial geometry represented by facial landmarks. Similar to its counterpart, the training process of $\rm{DLAT}_{lmk}$ also relies on four separate network components: the diverse age mapper $M_{lmk}$, the landmarks generator $G_{lmk}$, the age latent code predictor $P_{lmk}$, and the landmarks discriminator $D_{lmk}$. Their correlations during training are given in Fig. \ref{pipeline}(b).
\par

The function and implementation of $M_{lmk}$ completely inherit from $M_{img}$. Regarding the landmarks generator $G_{lmk}$, it is used to predict offsets caused by age changes with respect to the original landmarks on the input face. To be more concrete, it takes dimension-reduced facial landmarks, achieved by applying principle component analysis (PCA) to the original data in advance, as one of its inputs. Accordingly, the deformed landmarks can be obtained after applying inverse PCA to the generator's output estimation. The reason for these special designs is that we empirically found them favorable for generating stable results. More analysis about this will be given in Section \ref{ablation}. The age latent code is another input signal for $G_{lmk}$. We use age codes to modulate landmark-related features through FiLM operations \cite{perez2018film}. Since the inputs and outputs of $G_{lmk}$, $P_{lmk}$ and $D_{lmk}$ are all vector-formed representations of face landmarks, their network implementations are simple multi-layer perceptrons (MLPs). Relevant objectives, i.e., $\mathcal{L}_{div}^{lmk}$, $\mathcal{L}_{age}^{lmk}$ and $\mathcal{L}_{adv}^{lmk}$, are easy to deduce by replacing the image input with PCA-processed landmarks in their respective counterparts.
\par

\subsection{Consistency Constraints of Race and Pose}
Experiments show that simply encouraging the model to pursue diversity in age transformation usually leads to unreasonable results. For instance, given an input face with white skin, the $\rm{DLAT}_{img}$ occasionally generates age-transformed faces with black skin. Fed by a set of frontal facial landmarks, the $\rm{DLAT}_{lmk}$ sometimes yields deformed landmarks watching towards the sides. All these abnormal syntheses are contrary to our expectations: diverse and reasonable age transformation effects without modifying other facial attributes. In order to avoid unreasonable diversity, we propose race and pose consistency constraints for two age transformation systems, respectively.
\par 

The race consistency constraint for $\rm{DLAT}_{img}$ can be written as,
\begin{equation}
    \mathcal{L}_{rac} = \begin{cases}
        0,\quad \text{if}\ R(\boldsymbol{X})=R(\boldsymbol{Y}) \\
        \mathbb{E}_{\boldsymbol{X},\boldsymbol{Y}}(1-c), \quad \text{else}
    \end{cases},
\end{equation}
where $R$ is a third-party race estimator \cite{karkkainen2021fairface} that predicts a confidence (also called probability) distribution of the input face belonging to the considered race list. Here, $R(\boldsymbol{X})=R(\boldsymbol{Y})$ means the source and generated faces are classified into the same race class, and $c$ denotes the estimated confidence of the generated face categorized into the particular race class of the source face. Through minimizing $\mathcal{L}_{rac}$, the $\rm{DLAT}_{img}$ is forced to produce synthetic faces having the same ethnicity as the source. On the other hand, the plain philosophy behind our pose consistency constraint is to ensure that the poses of the input facial landmarks and deformed ones generated by the $\rm{DLAT}_{lmk}$ model are identical. We first construct a reference set composed of fixed and separated 3D points to obtain the pose approximation of those landmarks. Note that these points can be randomly chosen while satisfying the requirements of a pre-defined amount and mutual separability. Then, we pick an equal number of landmark pairs with the same locations on the face structure (e.g., the tip of the nose) from real and synthetic facial landmarks, constituting relevant landmark subsets. Finally, we compute the transformation from those 3D reference points to 2D selected landmarks with mature Perspective-n-Point (PnP) algorithms and regard it as the pose approximation corresponding to the current type of landmarks. The pose consistency restriction can then be formulated as follows:
\begin{equation}
\mathcal{L}_{pos}=\mathbb{E}_{\boldsymbol{L}_{\boldsymbol{X}}, \boldsymbol{L}_{\boldsymbol{Y}}}[||P(\boldsymbol{L}_{\boldsymbol{X}})-P(\boldsymbol{L}_{\boldsymbol{Y}})||],
\end{equation}
where $P(L_{X})$ and $P(L_{Y})$ denote pose calculation on real and generated facial landmarks, respectively.
\par

\subsection{Training Objectives}
\label{losses}
\textit{Compound identity preservation.} The original and synthesized faces need to seem like the same person perceptually, but only with differences in aging patterns. To achieve this goal, we design compound identity preservation losses to regularize the training processes of two age transformation models. Specifically, we employ three objectives for the $\rm{DLAT}_{img}$ to make regularization at pixel and feature levels. The first one is image cycle consistency, which is formulated as,
\begin{equation}
    \mathcal{L}_{cyc}^{img}=\mathbb{E}_{{\boldsymbol{X}}, {\boldsymbol{Y}}}[||\boldsymbol{X}-G_{img}(\boldsymbol{Y}, P_{img}(\boldsymbol{X}))||].
\end{equation}
The second one is perceptual consistency \cite{johnson2016perceptual}, written as,
\begin{equation}
    \mathcal{L}_{ppc}=\mathbb{E}_{{\boldsymbol{X}}, {\boldsymbol{Y}}}[||F_{vgg}(\boldsymbol{X})-F_{vgg}(G_{img}(\boldsymbol{Y}, P_{img}(\boldsymbol{X})))||],
\end{equation}
where $F_{vgg}$ computes feature maps from the first four layers of a pre-trained VGG \cite{simonyan2015deep}. The third one is identity feature consistency, which is defined as,
\begin{equation}
    \mathcal{L}_{idc}=\mathbb{E}_{{\boldsymbol{X}}, {\boldsymbol{Y}}}[1-cos(E_{fr}(\boldsymbol{X}), E_{fr}(\boldsymbol{Y}))],
\end{equation}
where $E_{fr}$ denotes using a pre-trained face recognition network (arcface \cite{deng2019arcface} in our implementation) to calculate identity-related feature vectors with unit norm lengths, and $cos(\cdot,\cdot)$ means calculating cosine similarity between the input vector pair.
\par

We devise two objectives for the $\rm{DLAT}_{lmk}$ model to preserve identity information. The first loss is the landmark cycle consistency $\mathcal{L}_{cyc}^{lmk}$, which has the same form as the $\mathcal{L}_{cyc}^{img}$. The other loss is personal characteristic consistency, which is defined as,
\begin{equation}
    \mathcal{L}_{pcc}=\mathbb{E}_{\boldsymbol{L}_{\boldsymbol{X}}, \boldsymbol{L}_{\boldsymbol{Y}}}[1-cos(\boldsymbol{L}_{\boldsymbol{X}}-\bar{\boldsymbol{L}^{r}}, \boldsymbol{L}_{\boldsymbol{Y}}-\bar{\boldsymbol{L}^{t}})].
\end{equation}
In this formula, $\boldsymbol{L}_{\boldsymbol{X}}$ and $\boldsymbol{L}_{\boldsymbol{Y}}$ are genuine and generated landmarks, and $\bar{\boldsymbol{L}^{r}}$ and $\bar{\boldsymbol{L}^{t}}$ are the averages of all real landmarks from the training datasets belonging to the source and target age groups, respectively. Through minimizing $\mathcal{L}_{pcc}$, the $\rm{DLAT}_{lmk}$ network is demanded to maintain the personal characteristics of the human subject in facial geometry. The reasonability behind this loss is straightforward. For example, if a child is born with a pair of bigger eyes than most children, this characteristic will remain for the rest of the corresponding subject's life.
\par

\textit{Full objective function.} For $\rm{DLAT}_{img}$, the whole losses used to supervise the model's training are:
\begin{equation}
\begin{split}
\mathcal{L}_{img}=&\lambda_{1}\mathcal{L}_{adv}^{img}-\lambda_{2}\mathcal{L}_{div}^{img}+\lambda_{3}\mathcal{L}_{age}^{img}+\lambda_{4}\mathcal{L}_{cyc}^{img}+\\
&\lambda_{5}\mathcal{L}_{rac}+\lambda_{6}\mathcal{L}_{ppc}+\lambda_{7}\mathcal{L}_{idc}.
\end{split}
\end{equation}
On the other hand, total objectives for $\rm{DLAT}_{lmk}$ can be summarized as:
\begin{equation}
\begin{split}
\mathcal{L}_{lmk}=&\lambda_{8}\mathcal{L}_{adv}^{lmk}-\lambda_{9}\mathcal{L}_{div}^{lmk}+\lambda_{10}\mathcal{L}_{age}^{lmk}+\lambda_{11}\mathcal{L}_{cyc}^{lmk}+\\
&\lambda_{12}\mathcal{L}_{pos}+\lambda_{13}\mathcal{L}_{pcc}.
\end{split}
\end{equation}
Note that these two networks are learned separately during the training stage. When it goes into the inference stage, $\rm{DLAT}_{img}$ can work independently or integrate with $\rm{DLAT}_{lmk}$ through image warping operations, which forms our enhanced diverse and lifespan age transformation model, $\rm{DLAT}^{\boldsymbol{+}}$. Regarding the image warping \cite{wolberg1990digital}, it is implemented as follows: with the original facial landmarks and their corresponding generated ones, we estimate the most possible transformation matrix between them by the least square method. After that, the same transform will be applied to the facial images produced by $\rm{DLAT}_{img}$, yielding the final age-shifted faces.

\section{Metric for Identity Deviation under Age Gaps}
\begin{figure}[tbp]
\centering
    \makebox[0.10\textwidth]{\scriptsize Input (0-2)}
    \makebox[0.10\textwidth]{\scriptsize Synthesis (30-39)}
    \makebox[0.10\textwidth]{\scriptsize Synthesis (50-69)}
    \vspace{2pt}
    \\
    \includegraphics[width=0.10\textwidth]{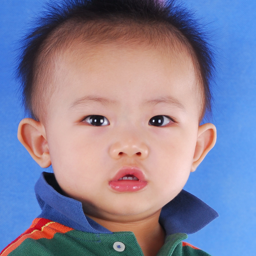}
    \includegraphics[width=0.10\textwidth]{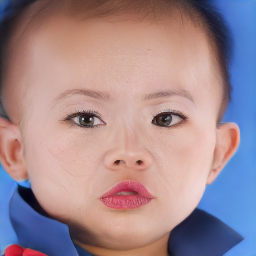}
    \includegraphics[width=0.10\textwidth]{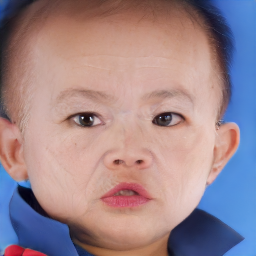}
    \\
    \makebox[0.10\textwidth]{\scriptsize (a)}
    \makebox[0.10\textwidth]{\scriptsize (b)}
    \makebox[0.10\textwidth]{\scriptsize (c)}
    \\
    \caption{Examples of poor age transformation synthesis that apparently contradict universal aging rules on human faces. However, these generations are appreciated when simply considering high face verification rates indicate more reasonable preservation of identity information across ages. The verification rates are $91.27\%$ between (a) and (b), $91.58\%$ between (a) and (c), measured by Face++ API \cite{FACEplus:API}.}
    \label{absurd_age_trans}
\end{figure}
To achieve the quantitative performance of an age transformation method, an important aspect is evaluating its ability to preserve the subject's identity information appropriately. Previous works \cite{li2021continuous,hsu2022agetransgan,liu2019attribute,yang2019learning,li2023pluralistic,10458950} usually resort to a well-performed face verification model and use it to calculate the average similarity between original and age-transformed faces. Then, a higher mean similarity score is regarded as a better result. However, according to our common sense, the appearances of the same person in different age stages show variations: they look similar but not identical. More specifically, the perceptual similarity between two faces of a particular human subject decreases as their age gap increases. Therefore, the conventional metric for identity preservation in the age-transformation scenario is unsuitable. Another piece of evidence is given in Fig. \ref{absurd_age_trans}, where the synthetic faces only demonstrate textural aging patterns. Nevertheless, those two absurd age transformation results are appreciated under the blind awarding of high face verification rates.
\par

\begin{figure*}[htb]
    \centering
    \includegraphics[width=0.18\linewidth]{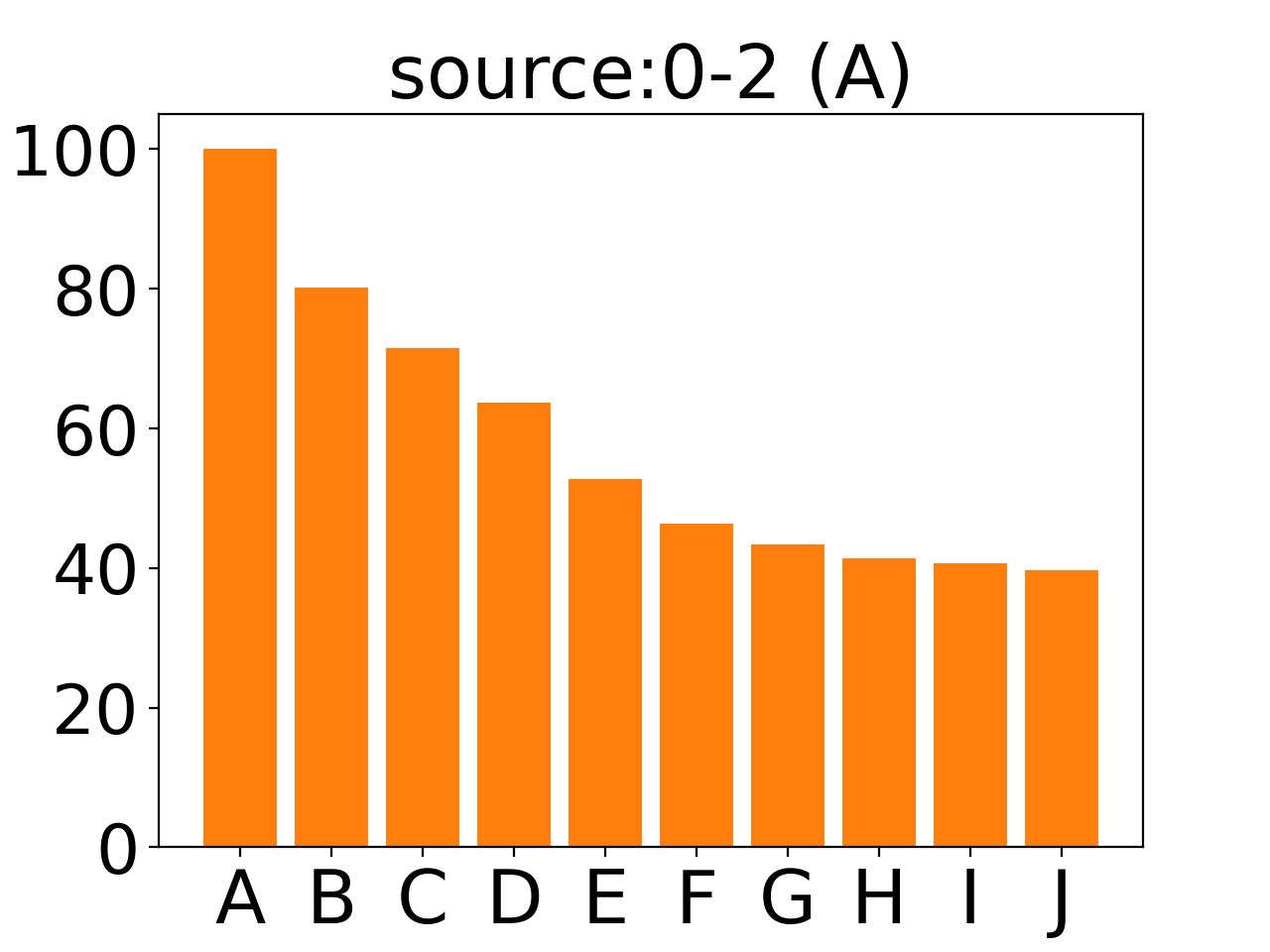}
    \includegraphics[width=0.18\linewidth]{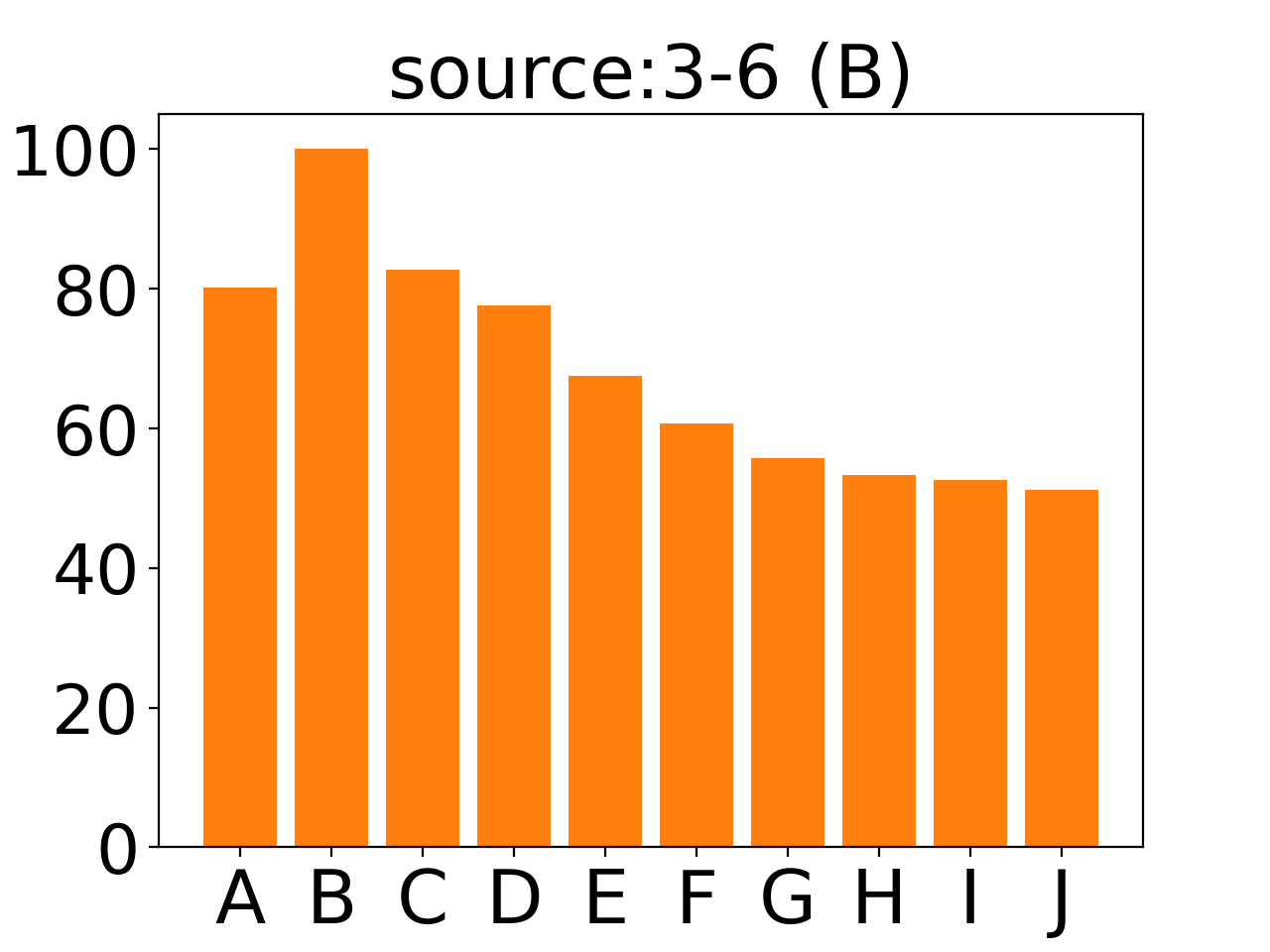}
    \includegraphics[width=0.18\linewidth]{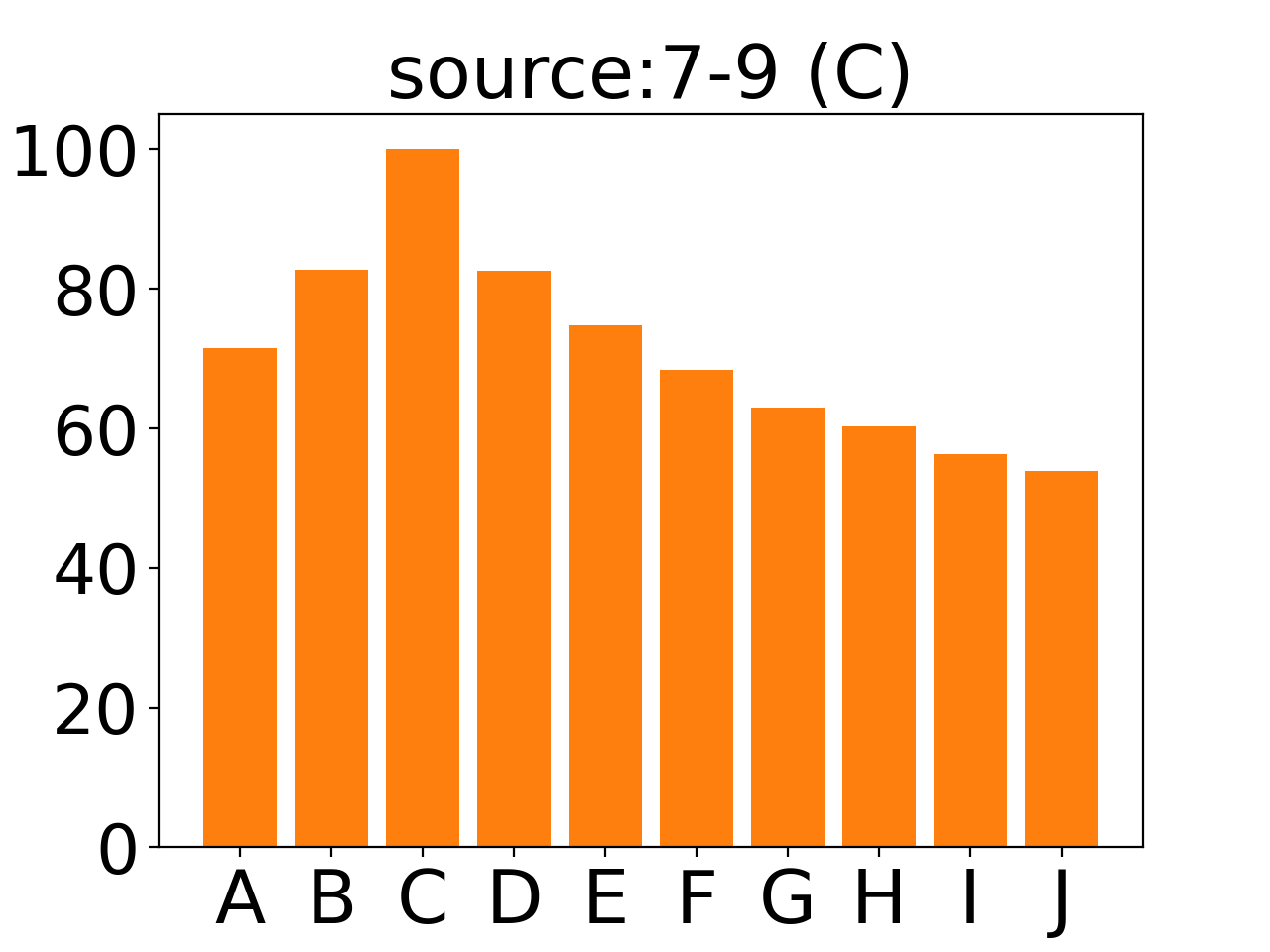}
    \includegraphics[width=0.18\linewidth]{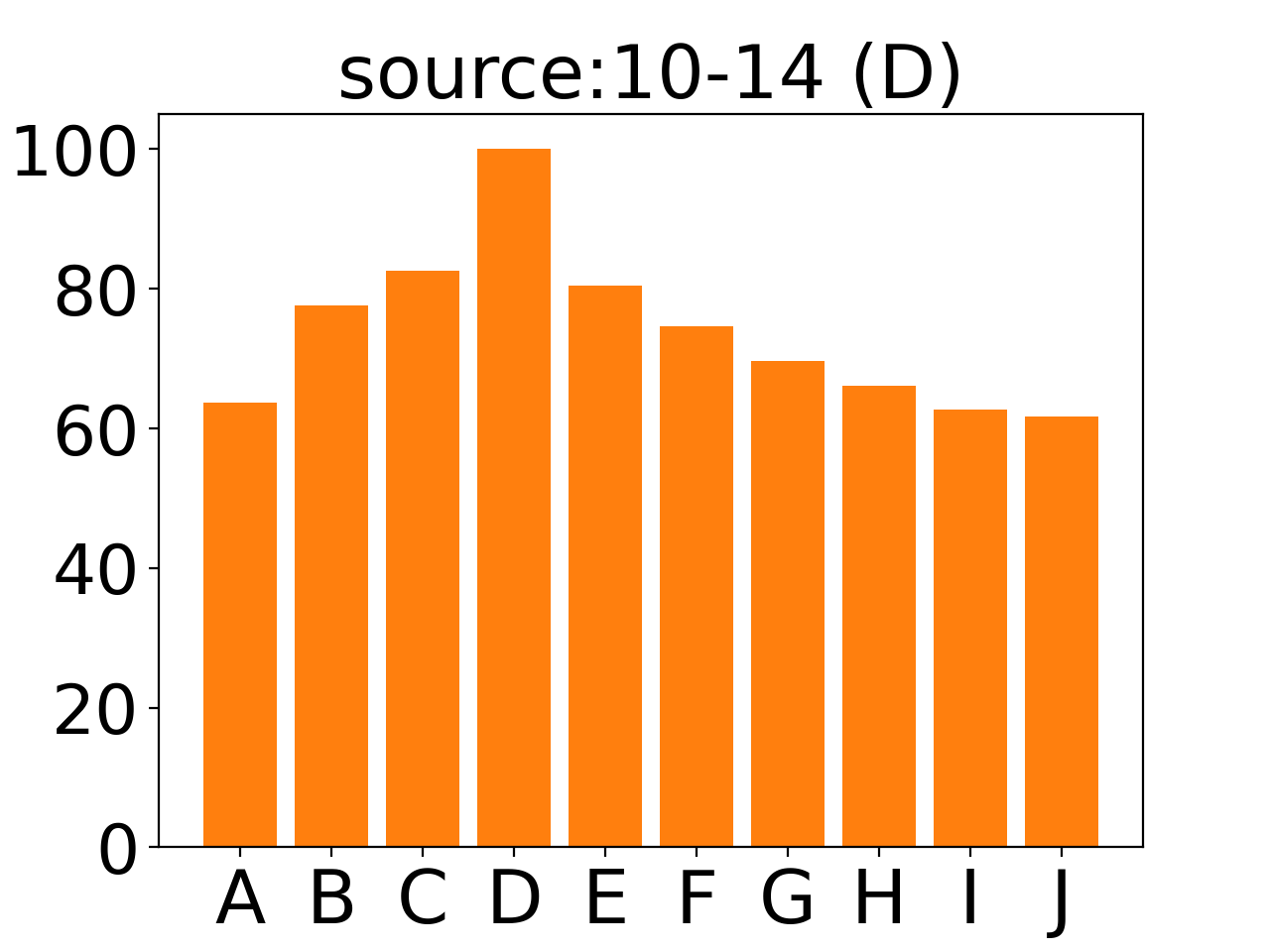}
    \includegraphics[width=0.18\linewidth]{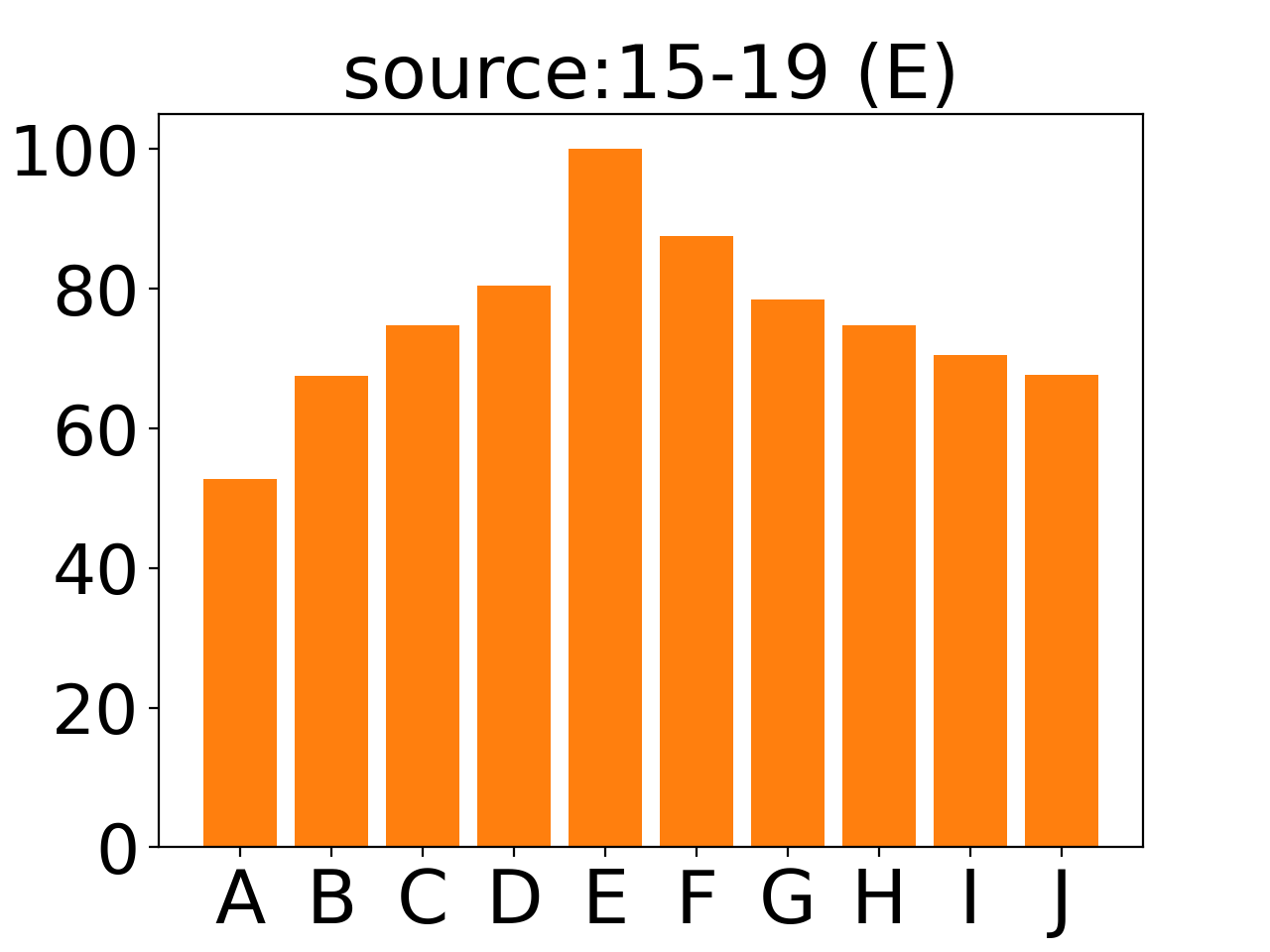}
    \vspace{5pt}
    \\
    \includegraphics[width=0.18\linewidth]{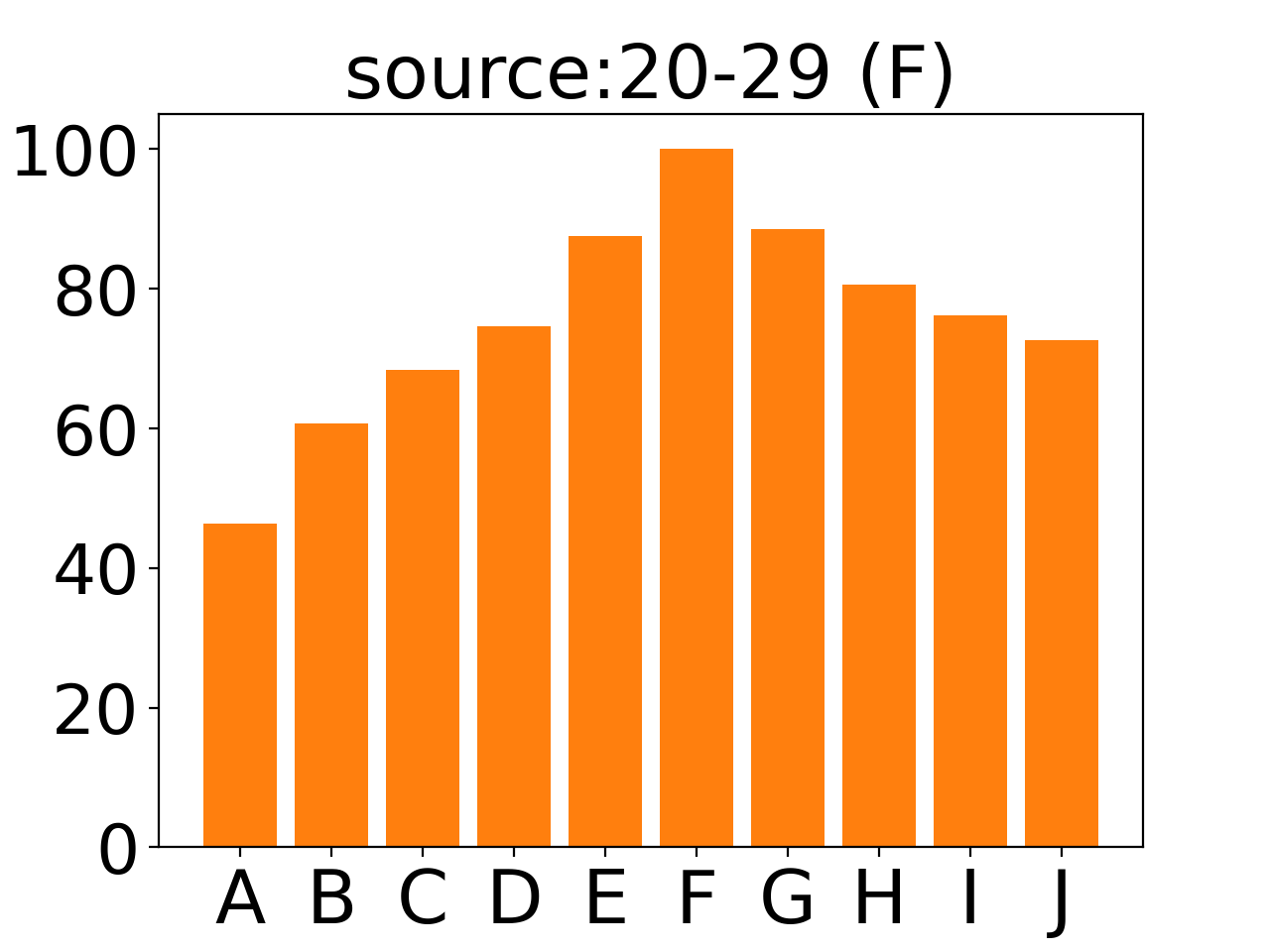}
    \includegraphics[width=0.18\linewidth]{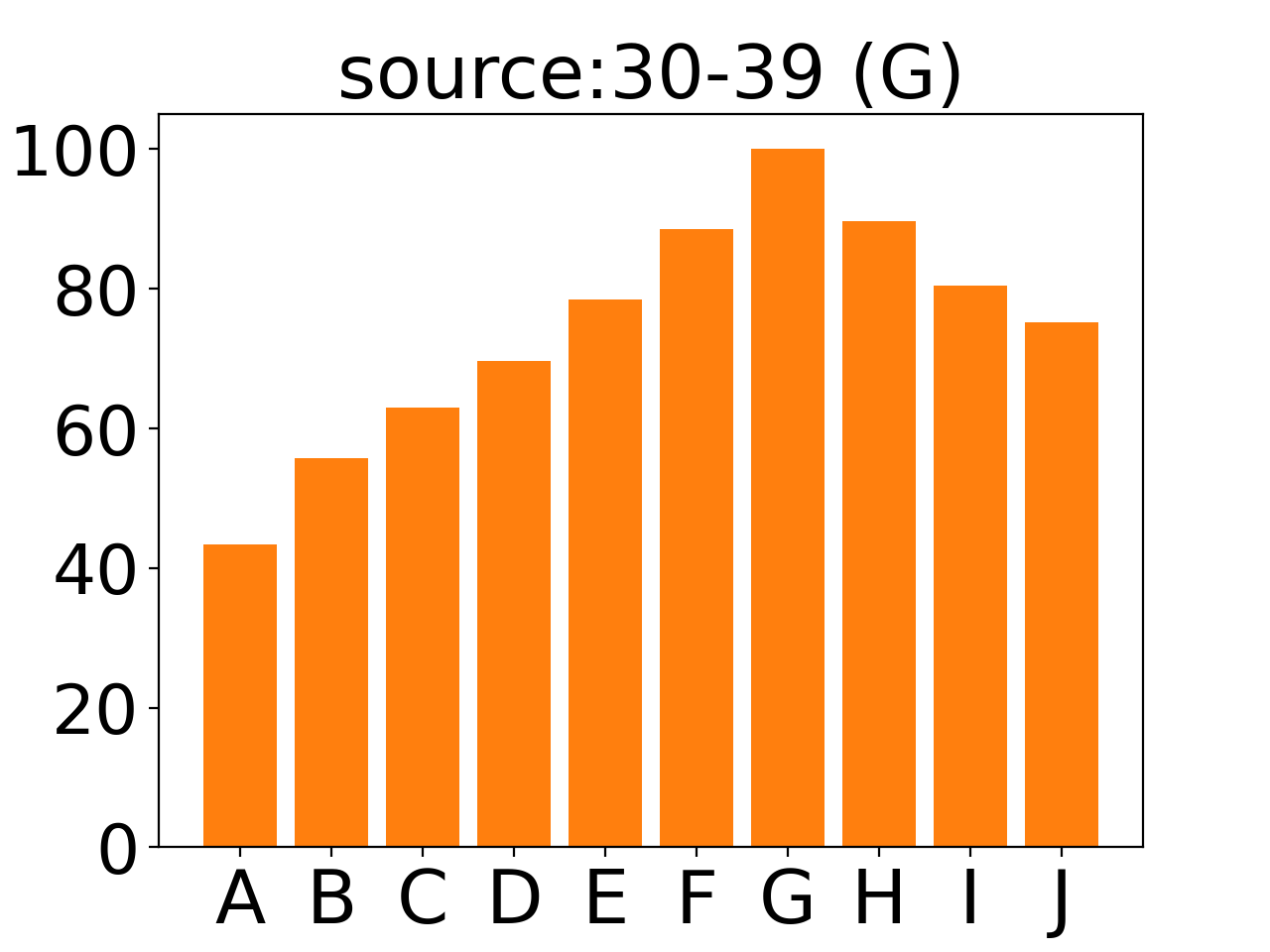}
    \includegraphics[width=0.18\linewidth]{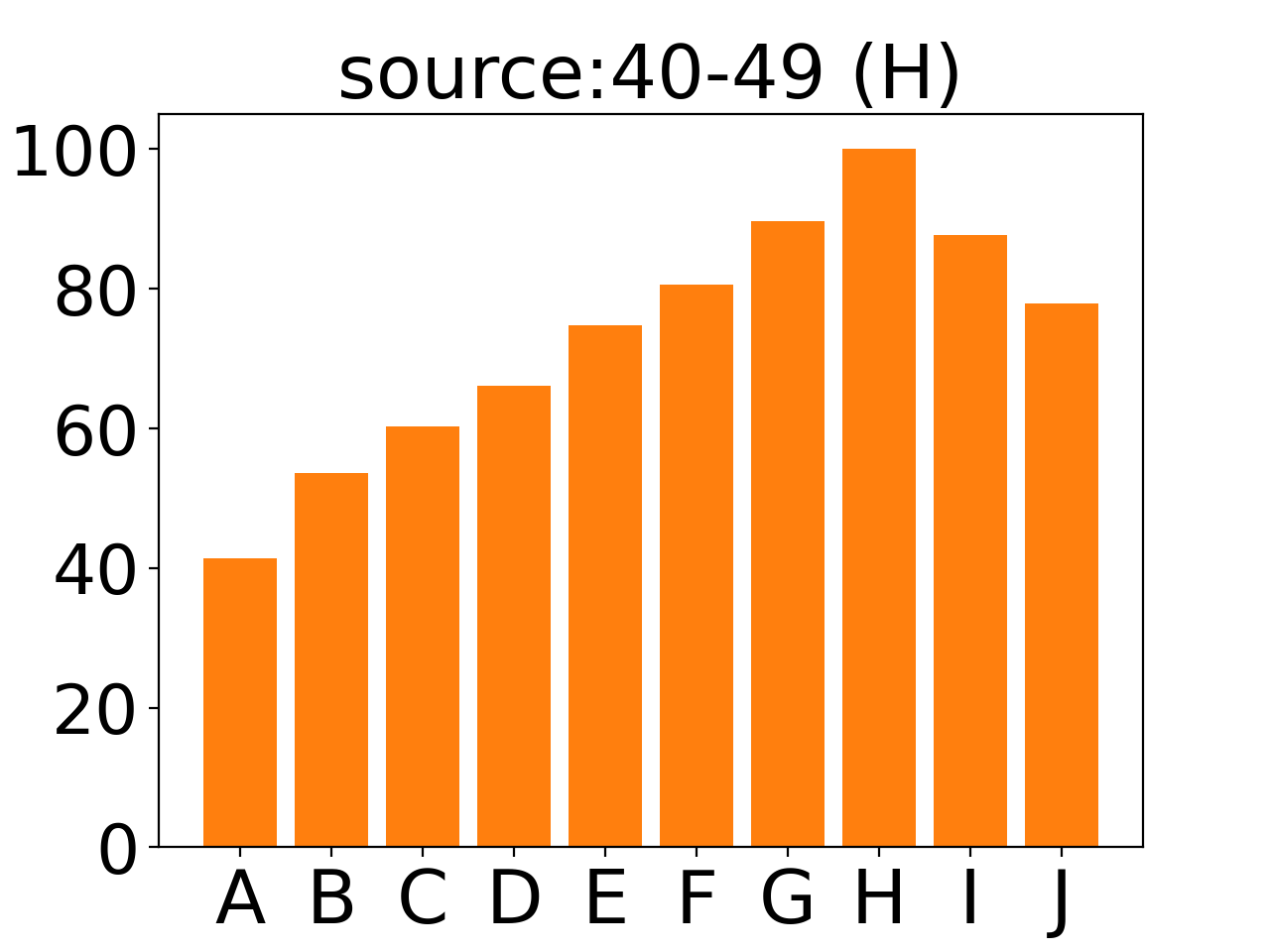}
    \includegraphics[width=0.18\linewidth]{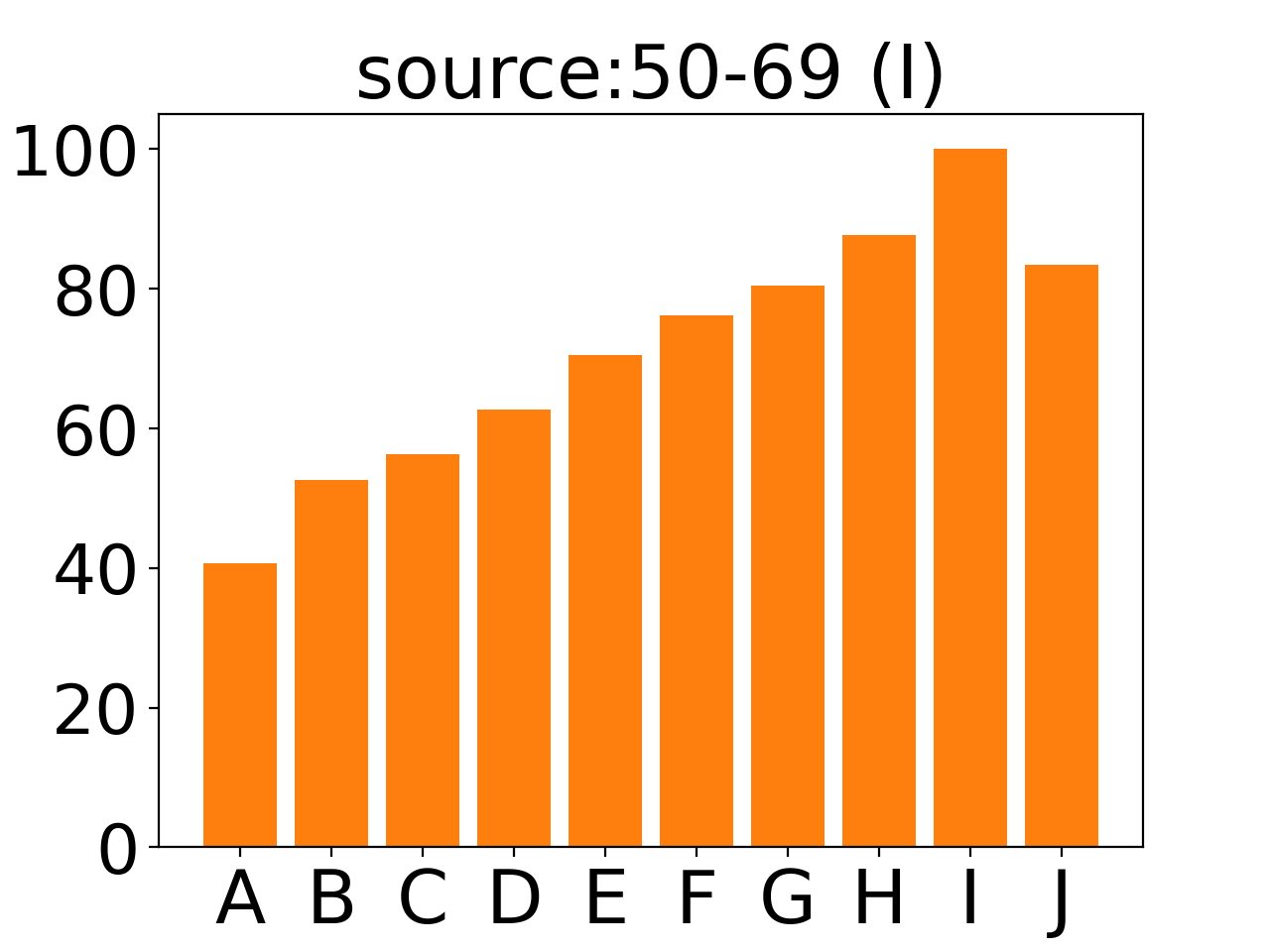}
    \includegraphics[width=0.18\linewidth]{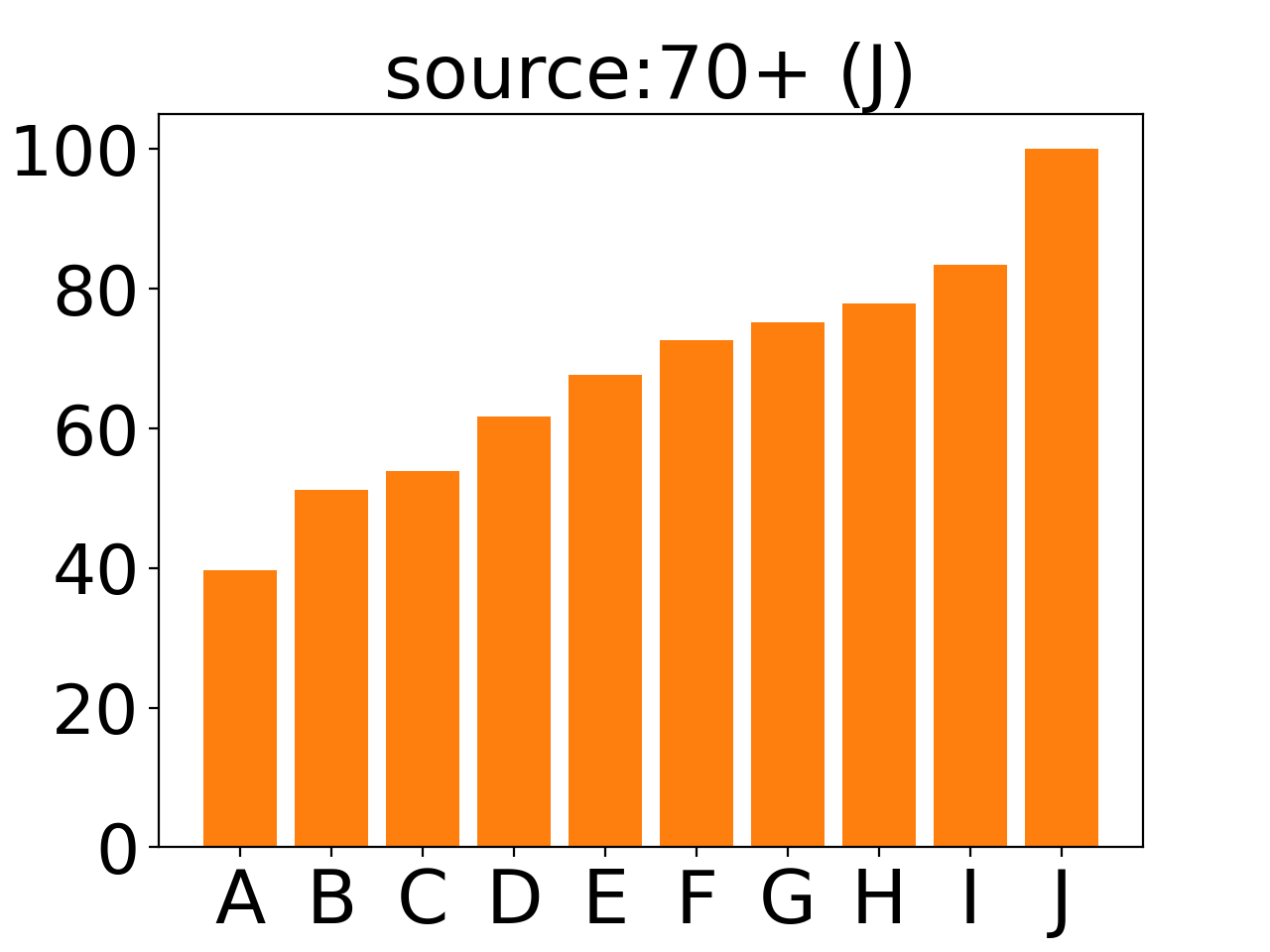}
    \caption{Statistical analysis on real face-aging datasets about average variation trends of identity information across ages. For each portrait of a specific age condition (the source age stage), we take it as the anchor and compare its identity similarities with faces from the same subject at identical or different phases. Similarity scores are calculated via the Face++ API. The symbols A to J denote ten consecutive age groups. These figures clearly show that identity loss becomes more severe as the increase of gap between the source and target age stages.
}
    \label{fig:our_metric}
\end{figure*}

In order to quantitatively investigate how the identity similarities of a face pair change with the varying age deviation between them in reality, we make analyses on the FG-NET \cite{Database:FGNET}, MORPH \uppercase\expandafter{\romannumeral2} \cite{ricanek2006morph}, AgeDB \cite{moschoglou2017agedb}, and Cross-Age Face \cite{hsu2022agetransgan} datasets, covering 13,617 subjects in total. \textit{It is worth noting that these four face datasets are the only public ones that provide subjects included with ground truth age labels and contain at least two portraits of the same person captured under different age stages. These conditions are necessary for the rationality of our analyses}. Following \cite{or2020lifespan}, we first divide the whole life cycle into ten consecutive and non-overlapped age groups, namely 0-2 (A), 3-6 (B), 7-9 (C), 10-14 (D), 15-19 (E), 20-29 (F), 30-39 (G), 40-49 (H), 50-69 (I), and 70+ (J). Then, for faces of each subject in these datasets, we use the face verification model provided by the Face++ API \cite{FACEplus:API} to calculate similarity scores of all image pairs. These scores are further categorized and averaged according to source and target age group pairs. Statistics for them summarized in Fig. \ref{fig:our_metric} clearly show the decreasing trends of calculated identity similarities between face pairs of the same person with increasing age differences, which aligns with universal face-aging rules observed in reality. For details of numerical values in these sub-figures, please refer to supplementary materials.

\par

With these ground-truth distributions of identity similarity, we propose our metric for \textbf{I}dentity \textbf{D}eviation under \textbf{A}ge \textbf{G}aps (IDAG), the definition of which is:
\begin{equation}
    IDAG = \frac{1}{N\times M}\sum_{i=1}^{N}\sum_{j=1}^{M}|\hat{s}(\boldsymbol{X}_i, \boldsymbol{Y}_i^j)-\bar{s}_i^j|\times100\%,
\end{equation}
where $N$ is the number of source faces involved and $M$ is the number of age groups considered (thus $1\leq M \leq 10$). In addition, $\hat{s}(\boldsymbol{X}_i, \boldsymbol{Y}_i^j)$ denotes the identity similarity predicted by the Face++ API between a face pair, the source one $\boldsymbol{X}_i$ and its corresponding synthesis $\boldsymbol{Y}_i^j$ in the predefined $j$th age phase. The notation $\bar{s}_i^j$ represents the mean similarity among real faces captured from the same persons under the source age group ($X_i$ belongs to) and the target age group ($\boldsymbol{Y}_i^j$ is located in), which has been computed and collected in Fig. \ref{fig:our_metric}. 
\par

\section{Experiments}
\subsection{Experimental Settings}
\textit{Datasets.} We use images from the FFHQ-Aging dataset \cite{Database:FFHQ-Aging} for models' training and evaluation purposes. The whole life duration is divided into ten separate age clusters based on the joint change of facial texture and shape at different age stages. However, only those clusters with sizes above a certain amount are chosen, considering the requirement of large numbers of training samples from the neural network. Eventually, there are 14,232 male and 14,066 female faces selected for training, 198 male and 205 female faces for testing. Facial images in the training set come from six age groups: 0-2, 3-6, 7-9, 15-19, 30-39, and 50-69. In addition, the FG-NET \cite{Database:FGNET}, MORPH \uppercase\expandafter{\romannumeral2} \cite{ricanek2006morph}, AgeDB \cite{moschoglou2017agedb}, and Cross-Age Face \cite{hsu2022agetransgan} datasets are leveraged to obtain the ground-truth mean similarity components of our IDAG metric.
\par

\textit{Baselines.} We take two kinds of methods into comparison. The first category includes state-of-the-art approaches particularly designed to deal with the lifespan age transformation task. They are LATS \cite{or2020lifespan}, DLFS \cite{he2021disentangled}, SAM \cite{alaluf2021only}, CUSP \cite{gomez2022custom}, and AgeTransGAN \cite{hsu2022agetransgan}. Methods in the second category contain InterFaceGAN \cite{shen2020interfacegan} and Latent-Transformer \cite{yao2021latent}, which are designed to realize general image attributes editing by manipulating relevant latent codes in the latent space of a StyleGAN2 \cite{karras2020analyzing} pretrained on the FFHQ \cite{Database:FFHQ} human face dataset. We employ the HyperStyle \cite{alaluf2022hyperstyle} for InterFaceGAN and pSp \cite{richardson2021encoding} for Latent-Transformer to obtain latent codes of source facial images. A noteworthy limitation of this type of image editing technique is that, although it can produce a younger or older synthesis of the input face, precise control of the target age condition is inaccessible. 
\par

\textit{Implementation Details.} Following the practices in \cite{he2021disentangled,or2020lifespan}, we train two separate models of our $\rm{DLAT}^{\boldsymbol{+}}$, one for males and the other for females. In addition, during the pre-processing stage, all images in the training set are resized to $256\times256$ resolution, and the background is removed using semantic masks provided by the dataset. An off-the-shelf detector \cite{lmkpredictor} is used to predict 81 landmarks for each given face. These 2D landmarks are further flattened to form a feature vector with a dimension of 162 and then processed by PCA to obtain the reduced version with a dimension of 32. All our models are learned on a single RTX 3090 GPU with a batch size of 2. We employ the Adam algorithm \cite{kingma2014adam} for parameter optimization, with an initial learning rate set to 0.001 and decayed by 0.1 at the $50th$ and $100th$ epochs, while the whole training lasts for 300 epochs. The length of age latent codes in $\rm{DLAT_{img}}$ and $\rm{DLAT_{lmk}}$ are set to 256 and 64, respectively. Regarding those hyper-parameters from $\lambda_{1}$ to $\lambda_{13}$ in Eqs. 11 and 12, we empirically assign them values of 1, 0.6, 10, 10, 1, 1, 0.2, 35, 2, 50, 10, 1, and 10.
\par

\textit{Evaluation Metrics.}  We evaluate methods for lifespan age transformation through automatic evaluation and user study. In automatic assessment, the proposed metric IDAG is used to measure the rationality of identity variation caused by age changes. Following the practices in \cite{choi2020stargan,liu2021divco}, the LPIPS \cite{zhang2018unreasonable} is calculated among a fixed number of synthesized faces in the same age cluster, where a higher value indicates a better diversity of face aging results. Besides, we employ the face detection API provided by Baidu AI Cloud \cite{baiduaicloud:API} to predict ages of synthetic faces and compute their absolute numerical differences with respect to corresponding target age groups we want to synthesize. These differences are summed up and then averaged, and the derived final result is referred to as the mean absolute error (MAE) in age transformation.
\par

\subsection{Quantitative Comparisons}
Since general image editing methods do not support age-related signals as the input, for the InterFaceGAN and Latent-Transformer involved, it is impossible to compute the age transformation error. However, we can use a well-performed age estimator to predict the corresponding age stages which their synthesized faces belong to. Furthermore, syntheses within the considered age clusters can be selected and compared with original input faces under the proposed IDAG metric. Quantitative results in terms of the identity deviation rationality and age transformation error yielded by our full model, $\rm{DLAT^{\boldsymbol{+}}}$, and other methods are tabulated in Tables \ref{IDAG_results} and \ref{ATE_results}. 
\par
\begin{table}[tbp]
\caption{Comparison of Identity Variation Rationality between Source and Synthetic Faces}
\centering
\begin{tabular}{c@{\hspace{7pt}}c@{\hspace{7pt}}c@{\hspace{7pt}}c@{\hspace{7pt}}c@{\hspace{7pt}}c@{\hspace{7pt}}c}
\toprule
\multirow{2}{*}{\diagbox[width=8em,trim=lr,outerleftsep=0pt, outerrightsep=2pt]{Method}{Source age}}     & 0-2    & 3-6    & 7-9    & 15-19 & 30-39  & 50-69  \\ 
\cline{2-7}
\xrowht{5pt}
& \multicolumn{6}{c}{IDAG$\boldsymbol{\downarrow}$} \\
\midrule  
LATS \cite{or2020lifespan} & 85.94  & 39.66  & 54.76  & 45.75 & 67.13  & 87.45  \\
DLFS \cite{he2021disentangled} & 62.09  & 39.42  & 49.02  & 42.76 & 61.56  & 56.00  \\
SAM \cite{alaluf2021only} & 141.93 & 88.44  & 83.44  & 69.74 & 92.60  & 90.10  \\ 
CUSP \cite{gomez2022custom} & 114.03 & 45.04  & 33.53  & \textbf{14.96} & \textbf{24.60}  & 31.35  \\
AgeTransGAN \cite{hsu2022agetransgan} & 175.37 & 122.79 & 110.96 & 92.00 & 110.22 & 128.17 \\
InterFaceGAN \cite{shen2020interfacegan} & 131.85 & 110.87 & 95.03  & 72.28 & 103.43  & 70.83  \\ 
Latent-Transformer \cite{yao2021latent} & 156.32 & 130.28  & 87.35  & 88.75 & 106.21  & 100.34  \\ 
$\rm{DLAT^{\boldsymbol{+}}}$ (\textbf{ours})        & \textbf{47.30}  & \textbf{29.83}  & \textbf{31.89}  & 18.80  & 29.20   & \textbf{29.36}  \\
\bottomrule
\end{tabular}
\label{IDAG_results}
\end{table}

\begin{table}[tbp]
\caption{Comparison of Age Transformation Error between Target and Synthetic Age Conditions}
\centering
\begin{tabular}{c@{\hspace{7pt}}c@{\hspace{7pt}}c@{\hspace{7pt}}c@{\hspace{7pt}}c@{\hspace{7pt}}c@{\hspace{7pt}}c}
\toprule
\multirow{2}{*}{\diagbox[width=8em,trim=lr,outerleftsep=0pt, outerrightsep=2pt]{Method}{Target age}}     & 0-2   & 3-6   & 7-9    & 15-19  & 30-39  & 50-69  \\ 
\cline{2-7}
\xrowht{5pt}
            & \multicolumn{6}{c}{MAE$\boldsymbol{\downarrow}$} \\
\midrule
LATS \cite{or2020lifespan} & 3.01 & 4.55 & 6.74  & 11.50 & 5.20  & 9.86  \\
DLFS \cite{he2021disentangled} & 2.29 & 4.57 & 6.27  & 10.19 & 5.03  & 10.50 \\
SAM \cite{alaluf2021only} & 6.20 & 5.83 & 6.86  & 8.08  & 4.81  & \textbf{6.66}  \\ 
CUSP \cite{gomez2022custom} & \textbf{0.84} & 2.73 & 5.44  & 8.57  & 18.93 & 11.90 \\
AgeTransGAN \cite{hsu2022agetransgan} & 5.95 & 8.22 & 11.16 & 11.96 & 9.06  & 17.57 \\
$\rm{DLAT^{\boldsymbol{+}}}$ (\textbf{ours})        & 1.41 & \textbf{2.59} & \textbf{5.11}  & \textbf{8.03}  & \textbf{4.50}  & 8.33  \\
\bottomrule
\end{tabular}
\label{ATE_results}
\end{table}

Results in table \ref{IDAG_results} show that: 1) Overall, it is rather difficult for current age transformation methods to fit the changing rules of human facial identity caused by age transition; 2) Approaches particularly designed for lifespan age transformation task demonstrate more reasonable variations of facial identity information across ages compared to those universal image editing techniques. Recall that the latter ones realize facial attribute modification by manipulating the corresponding latent codes in a pre-trained GAN space. Seeking to be capable of editing as many kinds of facial attributes as possible, this feature space eventually obtains a balance between feature disentanglement and editability. Because of the partial entanglement between identity and age attributes, the editing of facial age will incur the unwanted change of identity attribute; 3) In age clusters of 15-19 and 30-39, the performance of our method closely follows the best approaches. Remarkably, we surpass the second-best competitors by 23.8\%, 24,3\%, 4.9\%, and 6.3\% when the source age groups of input faces are 0-2, 3-6, 7-9, and 50-69, respectively. Note that when the input faces fall into those age ranges located at the start or end of the life cycle, the age gaps between them and corresponding syntheses covering the left age stages are larger than those cases that source faces belonging to the middle age groups. Results indicate that we can preserve the subject's identity more properly than others when the age transformation span is long. This superiority should be attributed to the compound losses proposed for identity preservation (see Section \ref{losses}).
\par

Table \ref{ATE_results} reveals that: 1) Given the six age groups as the synthesis target, it is easier for current age-changing algorithms to generate age-related patterns in early stages (0-9). However, it is more difficult to simulate old conditions. A possible reason for this phenomenon is that human faces widely undergo dramatic changes in facial shape during the early stage of life but relatively mild variations in texture after that period. Existing methods tend to be better at learning and modeling apparent differences than subtle ones; 2) our approach produces the lowest age transformation error in the middle four age ranges considered (3-6, 7-9, 15-19, and 30-39), outperforming the closest followers by 5.1\%, 6.1\%, 0.6\% and 6.4\%. While in the other two age groups, we also have competitive performance.  
\par

\begin{figure*}[tbp] 
\centering
    \makebox[0.10\textwidth]{\scriptsize Input (30-39)}
    \makebox[0.10\textwidth]{\scriptsize 0-2}
    \makebox[0.10\textwidth]{\scriptsize 3-6}
    \makebox[0.10\textwidth]{\scriptsize 7-9}
    \makebox[0.10\textwidth]{\scriptsize 15-19}
    \makebox[0.10\textwidth]{\scriptsize 30-39}
    \makebox[0.10\textwidth]{\scriptsize 50-69}
    \makebox[0.01\textwidth]{}
    \vspace{2pt}
    \\
    \includegraphics[width=0.10\textwidth]{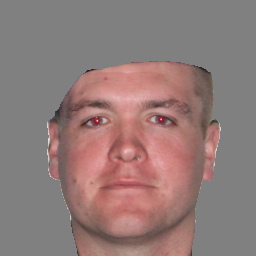}
    \includegraphics[width=0.10\textwidth]{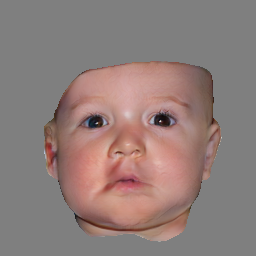}
    \includegraphics[width=0.10\textwidth]{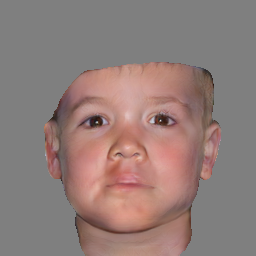}
    \includegraphics[width=0.10\textwidth]{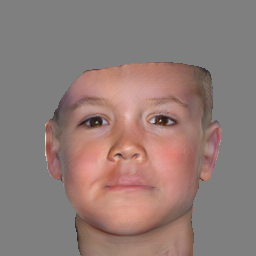}
    \includegraphics[width=0.10\textwidth]{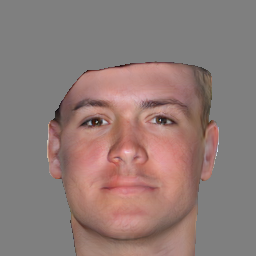}
    \includegraphics[width=0.10\textwidth]{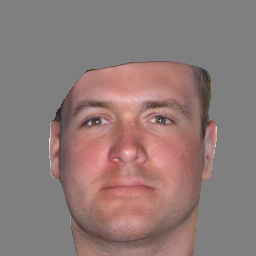}
    \includegraphics[width=0.10\textwidth]{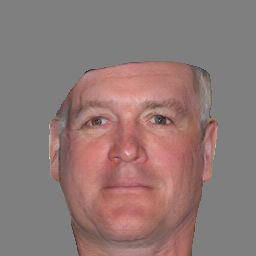}
    \raisebox{1.0\height}{\makebox[0.01\textwidth]{\rotatebox{90}{\makecell{\scriptsize LATS}}}}
    \vspace{3pt}
    \\
    \makebox[0.10\textwidth]{}
    \includegraphics[width=0.10\textwidth]{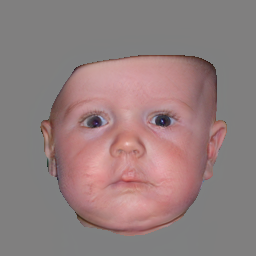}
    \includegraphics[width=0.10\textwidth]{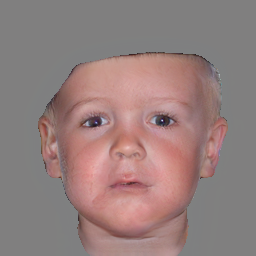}
    \includegraphics[width=0.10\textwidth]{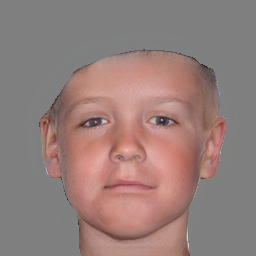}
    \includegraphics[width=0.10\textwidth]{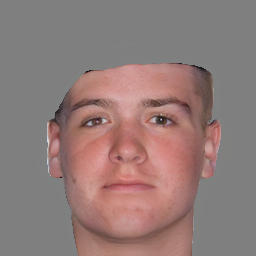}
    \includegraphics[width=0.10\textwidth]{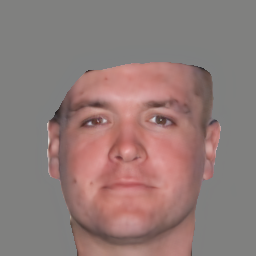}
    \includegraphics[width=0.10\textwidth]{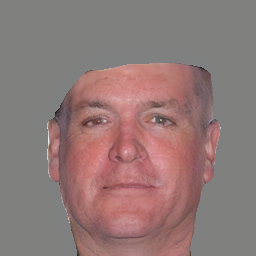}
    \raisebox{1.0\height}{\makebox[0.01\textwidth]{\rotatebox{90}{\makecell{\scriptsize DLFS}}}}
    \vspace{3pt}
    \\
    \makebox[0.10\textwidth]{}
    \includegraphics[width=0.10\textwidth]{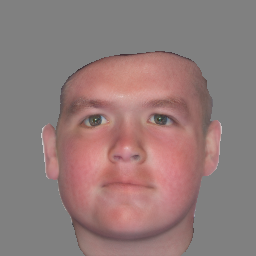}
    \includegraphics[width=0.10\textwidth]{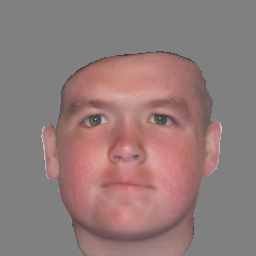}
    \includegraphics[width=0.10\textwidth]{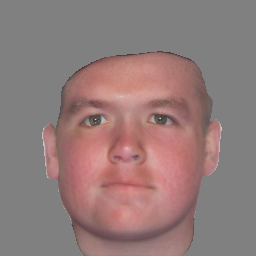}
    \includegraphics[width=0.10\textwidth]{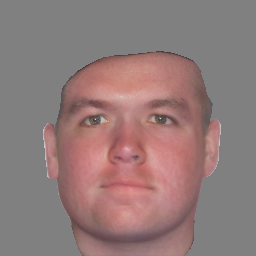}
    \includegraphics[width=0.10\textwidth]{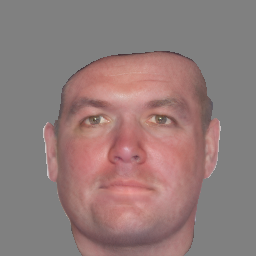}
    \includegraphics[width=0.10\textwidth]{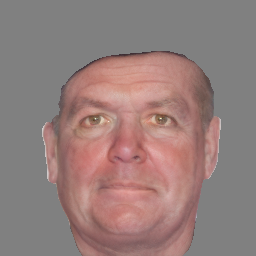}
    \raisebox{1.0\height}{\makebox[0.01\textwidth]{\rotatebox{90}{\makecell{\scriptsize SAM}}}}
    \vspace{3pt}
    \\
    \makebox[0.10\textwidth]{}
    \includegraphics[width=0.10\textwidth]{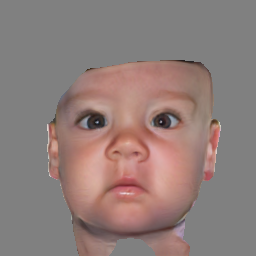}
    \includegraphics[width=0.10\textwidth]{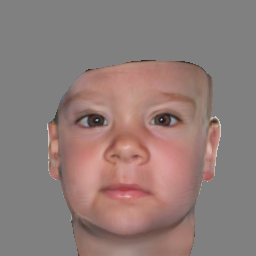}
    \includegraphics[width=0.10\textwidth]{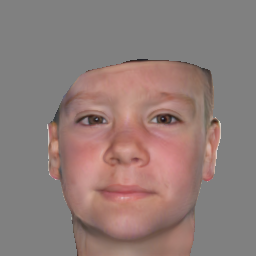}
    \includegraphics[width=0.10\textwidth]{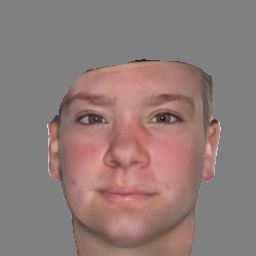}
    \includegraphics[width=0.10\textwidth]{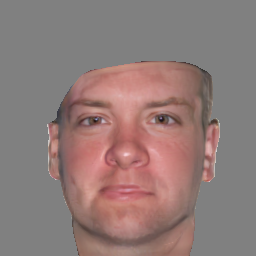}
    \includegraphics[width=0.10\textwidth]{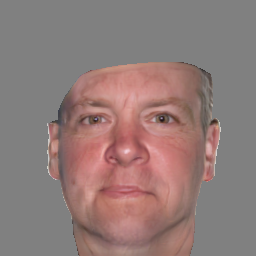}
    \raisebox{1.0\height}{\makebox[0.01\textwidth]{\rotatebox{90}{\makecell{\scriptsize CUSP}}}}
    \vspace{3pt}
    \\
    \makebox[0.10\textwidth]{}
    \includegraphics[width=0.10\textwidth]{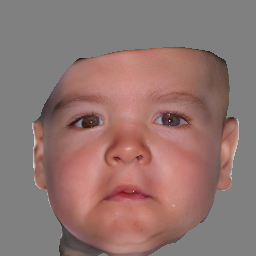}
    \includegraphics[width=0.10\textwidth]{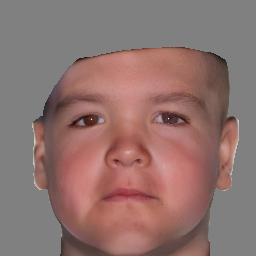}
    \includegraphics[width=0.10\textwidth]{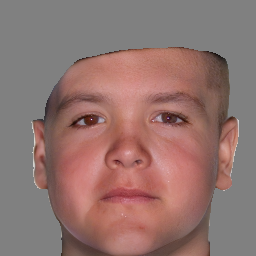}
    \includegraphics[width=0.10\textwidth]{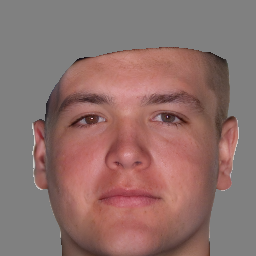}
    \includegraphics[width=0.10\textwidth]{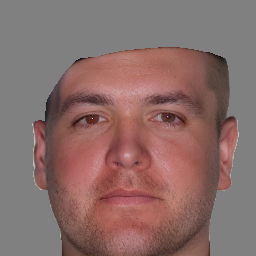}
    \includegraphics[width=0.10\textwidth]{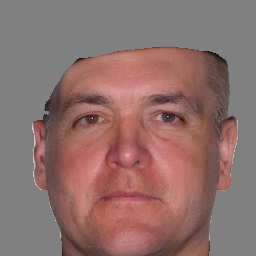}
    \raisebox{0.1\height}{\makebox[0.01\textwidth]{\rotatebox{90}{\makecell{\scriptsize AgeTransGAN}}}}
    \vspace{3pt}
    \\
    \makebox[0.10\textwidth]{}
    \includegraphics[width=0.10\textwidth]{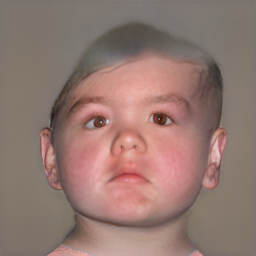}
    \includegraphics[width=0.10\textwidth]{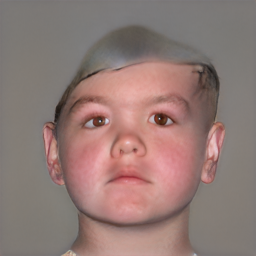}
    \includegraphics[width=0.10\textwidth]{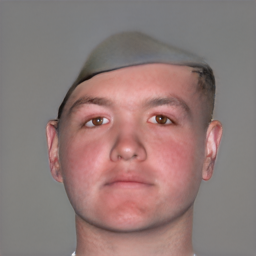}
    \includegraphics[width=0.10\textwidth]{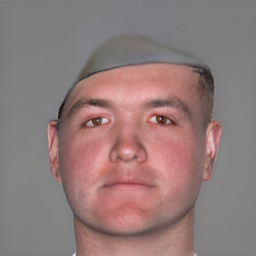}
    \includegraphics[width=0.10\textwidth]{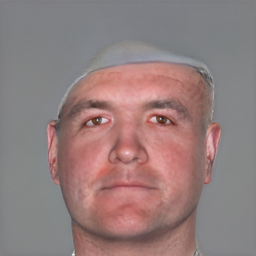}
    \includegraphics[width=0.10\textwidth]{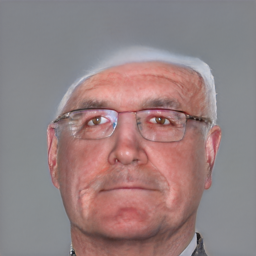}
    \raisebox{0.1\height}{\makebox[0.01\textwidth]{\rotatebox{90}{\makecell{\scriptsize InterFaceGAN}}}}
    \vspace{3pt}
    \\
    \makebox[0.10\textwidth]{}
    \includegraphics[width=0.10\textwidth]{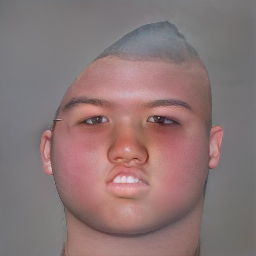}
    \includegraphics[width=0.10\textwidth]{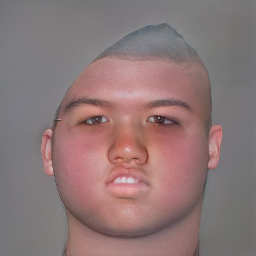}
    \includegraphics[width=0.10\textwidth]{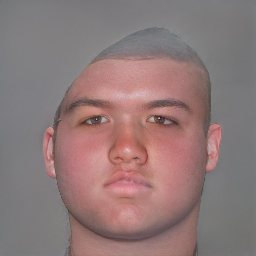}
    \includegraphics[width=0.10\textwidth]{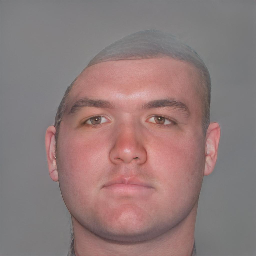}
    \includegraphics[width=0.10\textwidth]{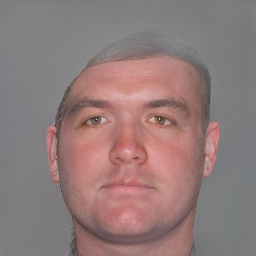}
    \includegraphics[width=0.10\textwidth]{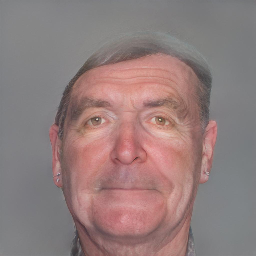}
    \raisebox{0.0\height}{\makebox[0.01\textwidth]{\rotatebox{90}{\makecell{\scriptsize Latent-Transformer}}}}
    \vspace{3pt}
    \\
    \makebox[0.10\textwidth]{}
    \includegraphics[width=0.10\textwidth]{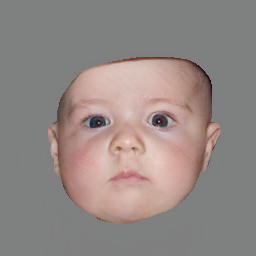}
    \includegraphics[width=0.10\textwidth]{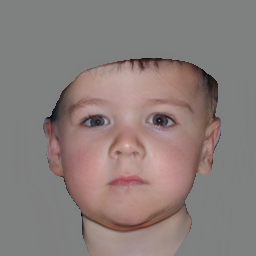}
    \includegraphics[width=0.10\textwidth]{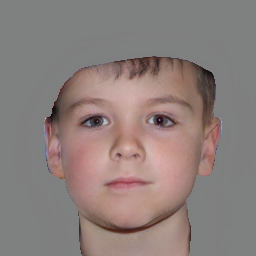}
    \includegraphics[width=0.10\textwidth]{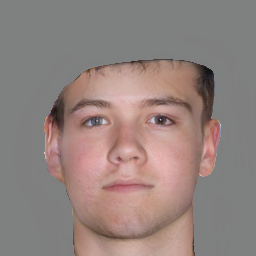}
    \includegraphics[width=0.10\textwidth]{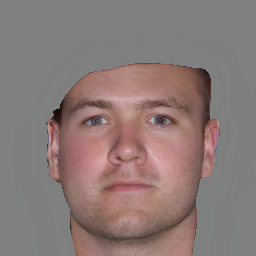}
    \includegraphics[width=0.10\textwidth]{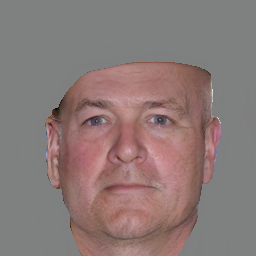}
    \raisebox{0.05\height}{\makebox[0.01\textwidth]{\rotatebox{90}{\makecell{\scriptsize $\rm{DLAT}^{\boldsymbol{+}}$ (\textbf{ours})}}}}
    \vspace{3pt}
    \\
    \caption{Visual comparison among lifespan age transformation results of the given male subject, which are synthesized by LATS, DLFS, SAM, CUSP, AgeTransGAN, InterFaceGAN, Latent-Transformer, and our $\rm{DLAT}^{\boldsymbol{+}}$.} 
    \label{fig:males}
\end{figure*}
\subsection{Qualitative and Subjective Comparisons}
Visualization of typical age-transformed results generated by all methods involved in comparison can be found in Fig. \ref{fig:males} and the supplementary material. These syntheses intuitively present the characteristics of each technique. In alignment with the quantitative evaluation results, approaches devised especially for facial age transformation (including LATS, DLFS, SAM, CUSP, AgeTransGAN, and ours) generally yield better-aging effects than common image editing methods. Checking the generations from InterFaceGAN and Latent-Transformer, we find several problems: 1) there are apparent changes in the subject's identity across ages; 2) Artifacts arise above the segmented facial regions where they should be empty, and unwanted ornaments (such as glasses) come out; 3) Unnatural facial syntheses occur when the target age conditions are located in childhood. We think the weak decoupling of features related to distinct facial attributes in the pre-trained GAN's latent space and the over-reliance on the prior knowledge embedded in this space jointly lead to the aforementioned issues. Drawbacks also exist among those specific algorithms designed for aging simulation. For instance, unexpected glasses appear in the synthetic female faces produced by LATS, as well as artifacts from CUSP and DLFS (see Fig. 12 in the appendix material). SAM usually generates abnormal texture changes (pay attention to the regions around the eyes). AgeTransGAN consistently produces nearly identical facial shapes across different ages, visually demonstrating its poor performance in identity variation, as reported in Table \ref{IDAG_results}. In contrast, faces synthesized by our method exhibit more reasonable aging patterns and consistent personal characteristics among all ages. Besides, our approach shows an interesting property: it can partially eliminate possible inappropriate illumination impacts on the captured faces (such as overexposure or underexposure) and restore the subject's original skin color in the corresponding facial syntheses with age progression or regression. We attribute this to the additional advantage of explicitly forcing race consistency between original and synthesized faces.
\par

On the other hand, we conduct subjective evaluations of our method and other compared approaches through the user study. To be more specific, 20 participants are asked to score age transformation syntheses generated by approaches in comparison. The subjective assessment involves six perspectives: \textit{identity consistency, shape deformation plausibility, texture transformation plausibility, synthetic image quality, age transformation error, and diversity}. Note that these aspects are mutually orthogonal. Statistics in Table \ref{usr_study_results} demonstrate that our algorithm achieves the best performance under all evaluation criteria, indicating a better alignment between human perceptual experience of face aging rules and our aging results.
\begin{table*}[tbp]
\caption{User Study of Age Transformation Results}
\centering
\begin{tabular}{ccccccc}
\toprule
Method & {\makecell[c]{Identity \\ consistency $\boldsymbol{\uparrow}$}}    &{\makecell[c]{Shape deformation \\ plausibility $\boldsymbol{\uparrow}$}}   & {\makecell[c]{Texture deformation \\ plausibility $\boldsymbol{\uparrow}$}}   & {\makecell[c]{Synthetic image \\ quality $\boldsymbol{\uparrow}$}}    & {\makecell[c]{Age transformation \\ error $\boldsymbol{\downarrow}$}}   & {\makecell[c]{Age transformation \\ diversity $\boldsymbol{\uparrow}$}}   \\ 
\midrule
LATS \cite{or2020lifespan} & 67.37 & 70.47 & 64.21 & 63.16 & 4.2  & 35.26 \\
DLFS \cite{he2021disentangled} & 73.16 & 73.68 & 76.84 & 72.63 & 3.4  & 33.68 \\
SAM \cite{alaluf2021only} & 54.21 & 40.87 & 51.05 & 61.58 & 17.3 & 34.74 \\
CUSP \cite{gomez2022custom} & 44.21 & 42.11 & 35.79 & 39.47 & 15.4 & 25.79 \\
AgeTransGAN \cite{hsu2022agetransgan} & 54.74 & 31.05 & 37.37 & 52.11 & 24.6 & 33.16 \\
InterFaceGAN \cite{shen2020interfacegan} & 50.00 & 57.37 & 62.11 & 60.00 & 23.2 & 38.95 \\
Latent-Transformer \cite{yao2021latent} & 32.10 & 35.79 & 34.21 & 31.58 & 7.1  & 31.58 \\ 
$\rm{DLAT^{\boldsymbol{+}}}$ (\textbf{ours}) & \textbf{83.68} & \textbf{79.85} & \textbf{79.47} & \textbf{78.37} & \textbf{2.6} & \textbf{74.73} \\
\bottomrule
\end{tabular}
\label{usr_study_results}
\end{table*}

\par

\subsection{Comparisons of Age Transformation Diversity}
\begin{figure*}[htbp]
\centering
    \makebox[0.10\textwidth]{\scriptsize Input (30-39)}
    \makebox[0.10\textwidth]{\scriptsize 0-2}
    \makebox[0.10\textwidth]{\scriptsize 3-6}
    \makebox[0.10\textwidth]{\scriptsize 7-9}
    \makebox[0.10\textwidth]{\scriptsize 15-19}
    \makebox[0.10\textwidth]{\scriptsize 30-39}
    \makebox[0.10\textwidth]{\scriptsize 50-69}
    \vspace{2pt}
    \\
    \includegraphics[width=0.10\textwidth]{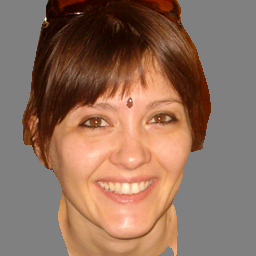}
    \includegraphics[width=0.10\textwidth]{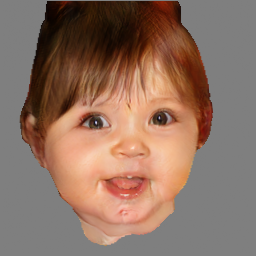}
    \includegraphics[width=0.10\textwidth]{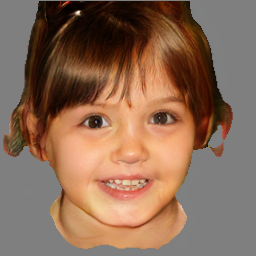}
    \includegraphics[width=0.10\textwidth]{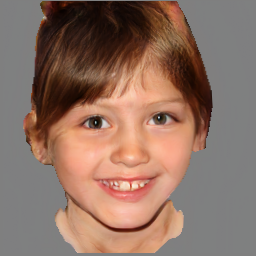}
    \includegraphics[width=0.10\textwidth]{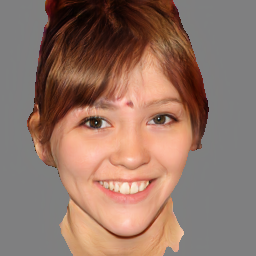}
    \includegraphics[width=0.10\textwidth]{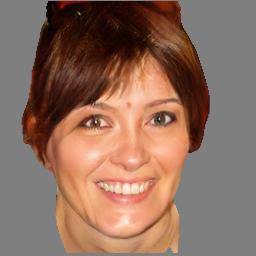}
    \includegraphics[width=0.10\textwidth]{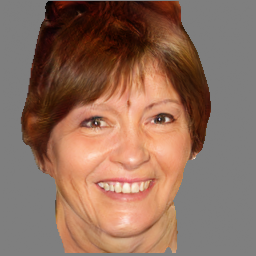}
    \vspace{3pt}
    \\
    \makebox[0.10\textwidth]{}
    \includegraphics[width=0.10\textwidth]{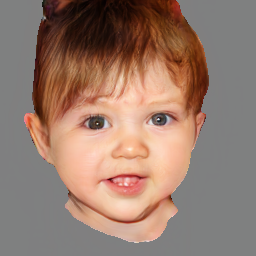}
    \includegraphics[width=0.10\textwidth]{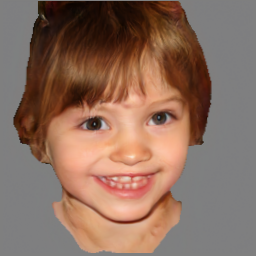}
    \includegraphics[width=0.10\textwidth]{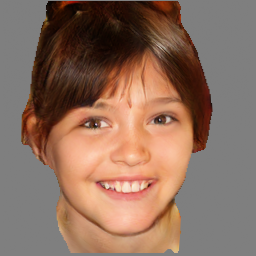}
    \includegraphics[width=0.10\textwidth]{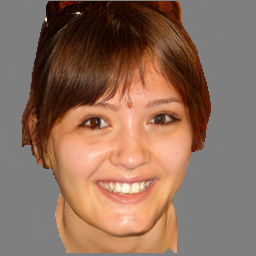}
    \includegraphics[width=0.10\textwidth]{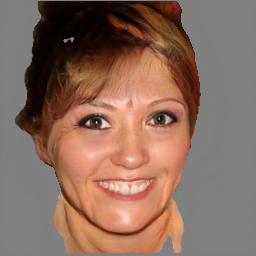}
    \includegraphics[width=0.10\textwidth]{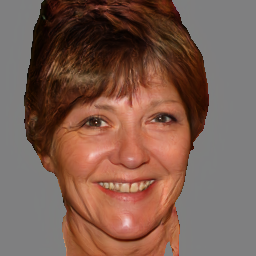}
    \vspace{3pt}
    \\
    \makebox[0.10\textwidth]{}
    \includegraphics[width=0.10\textwidth]{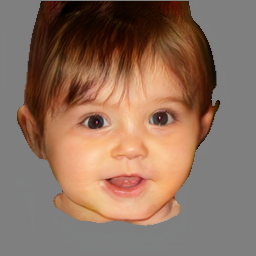}
    \includegraphics[width=0.10\textwidth]{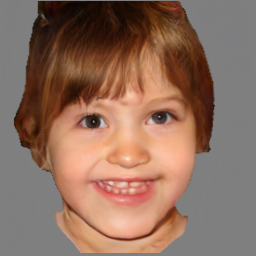}
    \includegraphics[width=0.10\textwidth]{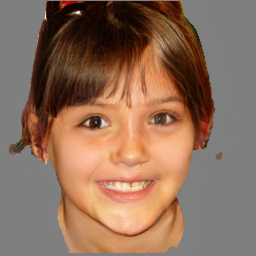}
    \includegraphics[width=0.10\textwidth]{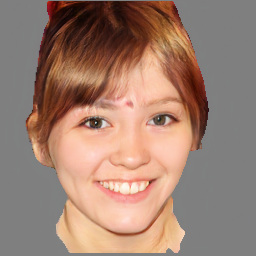}
    \includegraphics[width=0.10\textwidth]{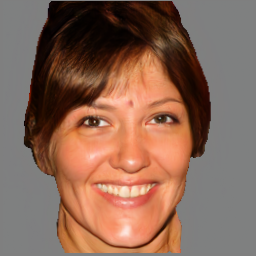}
    \includegraphics[width=0.10\textwidth]{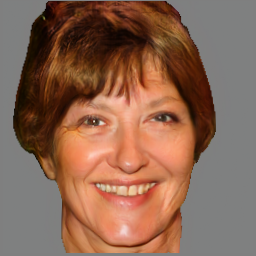}
    \vspace{3pt}
    \\
    \makebox[0.10\textwidth]{}
    \includegraphics[width=0.10\textwidth]{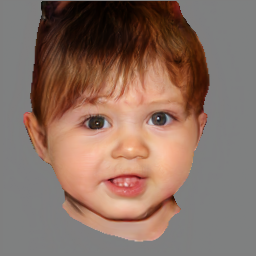}
    \includegraphics[width=0.10\textwidth]{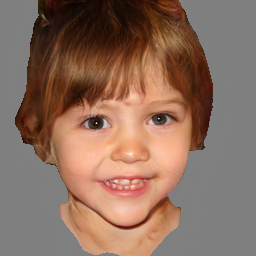}
    \includegraphics[width=0.10\textwidth]{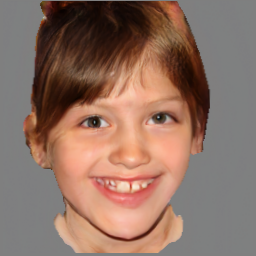}
    \includegraphics[width=0.10\textwidth]{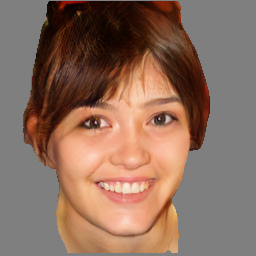}
    \includegraphics[width=0.10\textwidth]{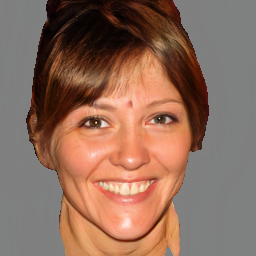}
    \includegraphics[width=0.10\textwidth]{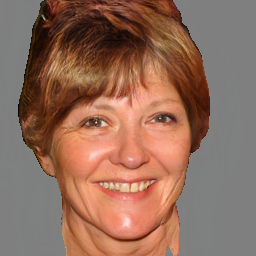}
    \vspace{3pt}
    \\
\caption{Visualization of diverse age transformation results across the life cycle, generated by applying our full method to a female subject.}
\label{fig:females_diversity}
\end{figure*}

Quantitative evaluations about the age transformation diversity of our method and other approaches are tabulated in Table \ref{synthesis-divesity-LPIPS}. The results show that our algorithm beats those opponents by several orders of magnitude under the LPIPS metric. After checking the synthesized faces under every specific target age group generated by those approaches, we could hardly find subtle variations among their corresponding generations. In other words, these methods only yield deterministic aging effects. However, according to common sense, a person's appearance at any age timestamp should have infinite possibilities. On the other hand, representative visualization results produced by our technique are shown in Fig. \ref{fig:females_diversity} and supplementary materials. These synthesized faces show that our full model, $\rm{DLAT}^{\boldsymbol{+}}$, is capable of generating photorealistic and vivid age-progressed or regressed faces of the input portrait. At the same time, the diversity is prominently reflected in facial texture and shape levels. This exclusive advantage compared to other age transformation methods should be attributed to our special design of the diverse mechanism in the age-conditioned synthetic procedure and the proposed facial geometry rectification scheme based on face landmarks, accompanied by multiple constraints on race, pose and identity consistency to control the aging diversity within a reasonable range.
\begin{table*}[tbp]
\caption{Comparison of Age Transformation Diversity under Certain Target Age Groups}
\centering
\renewcommand{\arraystretch}{1.1}
\begin{tabular}{ccccccc} 
\toprule
\multirow{2}{*}{\diagbox[width=8em,trim=lr,outerleftsep=0pt, outerrightsep=2pt]{Method}{Target age}} & 0-2 & 3-6 & 7-9 & 15-19 & 30-39 & 50-69 \\ 
\cline{2-7}
\xrowht{5pt}
& \multicolumn{6}{c}{LPIPS$\boldsymbol{\uparrow}$} \\
\midrule
LATS \cite{or2020lifespan} & \num{2.1e-2} & \num{1.7e-2} & \num{1.3e-2} & \num{1.3e-2}   & \num{1.4e-2}  & \num{1.5e-2}   \\
DLFS \cite{he2021disentangled} & \num{1.6e-2} & \num{1.5e-2} & \num{9.7e-3} & \num{7.3e-3}   & \num{5.6e-3} & \num{9.4e-3}   \\
SAM \cite{alaluf2021only} & \num{9.2e-5} & \num{9.2e-5} & \num{9.0e-5} & \num{8.8e-5}   & \num{8.9e-5} & \num{9.6e-5}   \\ 
CUSP \cite{gomez2022custom} & \num{2.0e-2} & \num{2.1e-2} & \num{2.2e-2} & \num{2.1e-2}   & \num{1.6e-2} & \num{2.0e-2}   \\
AgeTransGAN \cite{hsu2022agetransgan} & \num{1.7e-2} & \num{1.6e-2} & \num{1.3e-2}  & \num{1.8e-2} & \num{1.7e-2} & \num{1.8e-2}   \\
InterFaceGAN \cite{shen2020interfacegan} & \num{3.1e-2} & \num{2.7e-2}  & \num{3.4e-2}  & \num{2.3e-2} & \num{2.5e-2}  & \num{2.7e-2}  \\ 
Latent-Transformer \cite{yao2021latent} & \num{2.8e-2} & \num{3.2e-2}  & \num{2.3e-2}  & \num{2.8e-2} & \num{2.9e-2}  & \num{3.3e-2}  \\ 
$\rm{DLAT^{\boldsymbol{+}}}$ (\textbf{ours}) & \pmb{\num{2.0e-1}} & \pmb{\num{1.4e-1}} & \pmb{\num{1.3e-1}} & \pmb{\num{1.7e-1}} & \pmb{\num{1.8e-1}} & \pmb{\num{1.8e-1}}   \\
\bottomrule
\end{tabular}
\label{synthesis-divesity-LPIPS}
\end{table*}

\par

\subsection{Ablation Studies} \label{ablation}
\begin{figure*}[tp] 
\centering
    \makebox[0.10\textwidth]{\scriptsize Input (3-6)}
    \makebox[0.10\textwidth]{\scriptsize 0-2}
    \makebox[0.10\textwidth]{\scriptsize 3-6}
    \makebox[0.10\textwidth]{\scriptsize 7-9}
    \makebox[0.10\textwidth]{\scriptsize 15-19}
    \makebox[0.10\textwidth]{\scriptsize 30-39}
    \makebox[0.10\textwidth]{\scriptsize 50-69}
    \makebox[0.01\textwidth]{}
    \vspace{2pt}
    \\
    \includegraphics[width=0.10\textwidth]{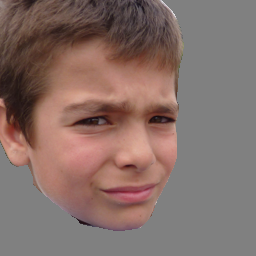}
    \includegraphics[width=0.10\textwidth]{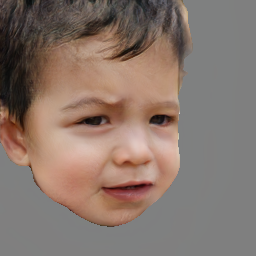}
    \includegraphics[width=0.10\textwidth]{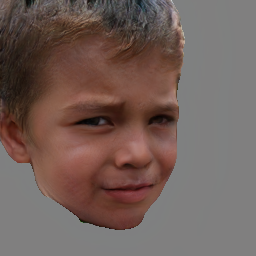}
    \includegraphics[width=0.10\textwidth]{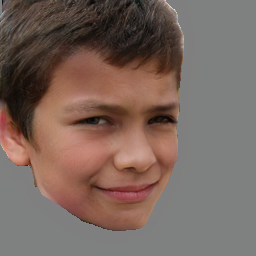}
    \includegraphics[width=0.10\textwidth]{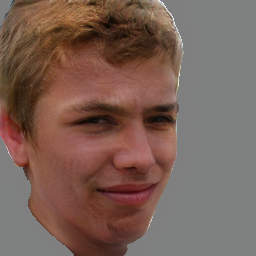}
    \includegraphics[width=0.10\textwidth]{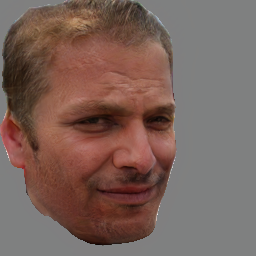}
    \includegraphics[width=0.10\textwidth]{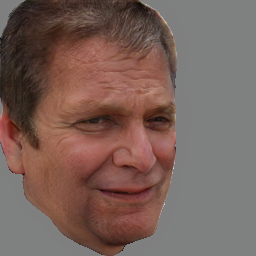}
    \raisebox{0.3\height}{\makebox[0.01\textwidth]{\rotatebox{90}{\makecell{\scriptsize w/o $\mathcal{L}_{rac}$}}}}
    \vspace{3pt}
    \\
    \makebox[0.10\textwidth]{}
    \includegraphics[width=0.10\textwidth]{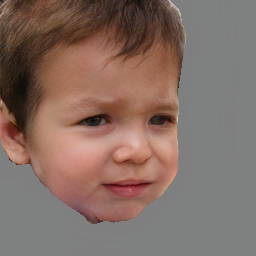}
    \includegraphics[width=0.10\textwidth]{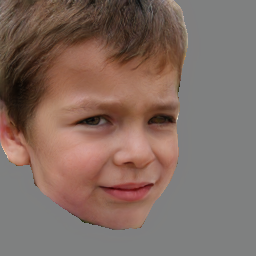}
    \includegraphics[width=0.10\textwidth]{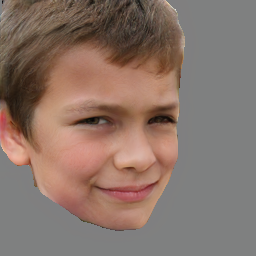}
    \includegraphics[width=0.10\textwidth]{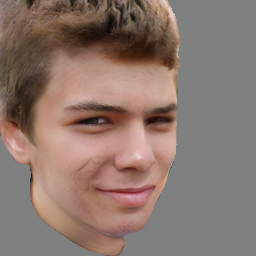}
    \includegraphics[width=0.10\textwidth]{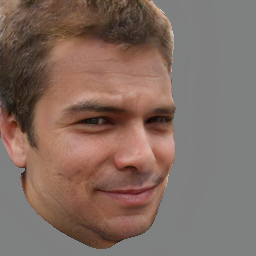}
    \includegraphics[width=0.10\textwidth]{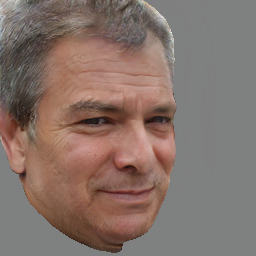}
    \raisebox{0.5\height}{\makebox[0.01\textwidth]{\rotatebox{90}{\makecell{\scriptsize w/ $\mathcal{L}_{rac}$}}}}
    \vspace{3pt}
    \\
    \caption{Generations comparison of training our $\rm{DLAT}_{img}$ without (w/o) or with (w/) the race consistency constraint.} 
    \label{fig:ablation_L_race}
\end{figure*}

\begin{figure}[tbp] 
    \centering
    \makebox[0.09\textwidth]{\scriptsize Input (50-69)}
    \makebox[0.09\textwidth]{\scriptsize 0-2}
    \makebox[0.09\textwidth]{\scriptsize 0-2}
    \makebox[0.09\textwidth]{\scriptsize 0-2}
    \makebox[0.09\textwidth]{\scriptsize 0-2}
    \\ \vspace{2pt}
    \includegraphics[width=0.09\textwidth]{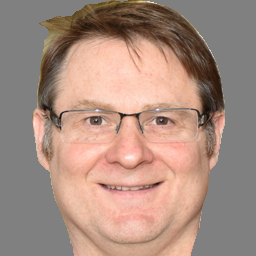}
    \includegraphics[width=0.09\textwidth]{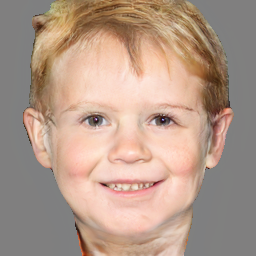}
    \includegraphics[width=0.09\textwidth]{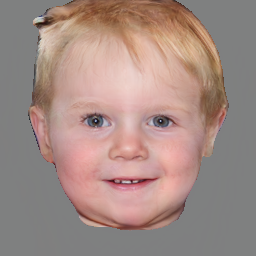}
    \includegraphics[width=0.09\textwidth]{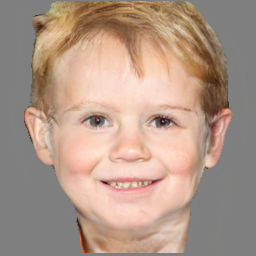}
    \includegraphics[width=0.09\textwidth]{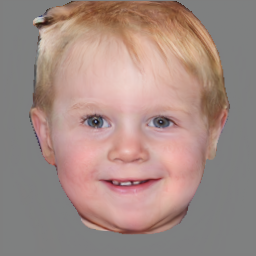}
    \\ \vspace{2pt}
    \makebox[0.09\textwidth]{}
    \includegraphics[width=0.09\textwidth]{ablation/full_model/baseline.png}
    \includegraphics[width=0.09\textwidth]{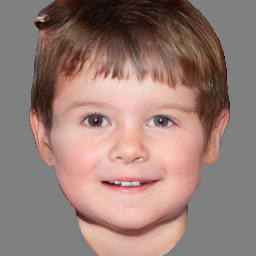}
    \includegraphics[width=0.09\textwidth]{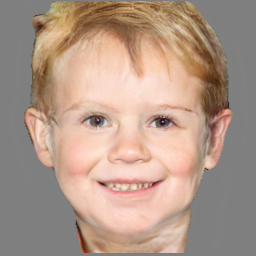}
    \includegraphics[width=0.09\textwidth]{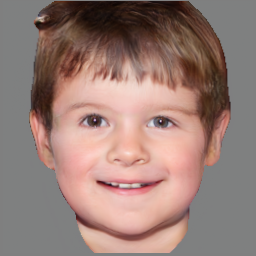}
    \\ \vspace{2pt}
    \makebox[0.09\textwidth]{}
    \includegraphics[width=0.09\textwidth]{ablation/full_model/baseline.png}
    \includegraphics[width=0.09\textwidth]{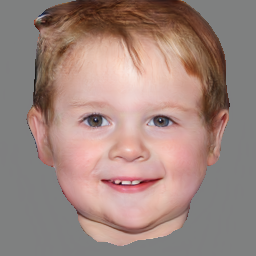}
    \includegraphics[width=0.09\textwidth]{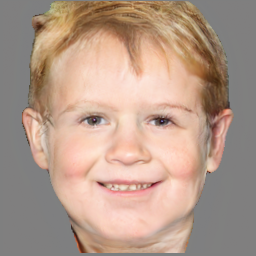}
    \includegraphics[width=0.09\textwidth]{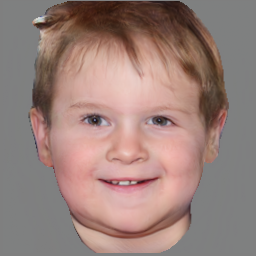}
    \\ 
    \makebox[0.09\textwidth]{}
    \makebox[0.09\textwidth]{\scriptsize V1}
    \makebox[0.09\textwidth]{\scriptsize V2}
    \makebox[0.09\textwidth]{\scriptsize V3}
    \makebox[0.09\textwidth]{\scriptsize Ours full}
    \caption{Visual comparison among results of transforming the input face to the age range of 0 to 2. From the second to the fifth column, each is connected to a specific variant of our approach. V2 denotes the $\rm{DLAT}_{img}$, while V1 is its variant through ablating the diversity mechanism. V3 represents the version of V1 plus $\rm{DLAT}_{lmk}$. The full model is the integration of $\rm{DLAT}_{img}$ and $\rm{DLAT}_{lmk}$.}
    \label{fig:ablation-global}
\end{figure}

\begin{table}[tbp]
\caption{Ablation Study of Diversity Mechanisms in Our Method}
\label{diversity_ablation}
\centering
\begin{tabular}{c@{\hspace{7pt}}c@{\hspace{7pt}}c@{\hspace{7pt}}c@{\hspace{7pt}}c@{\hspace{7pt}}c@{\hspace{7pt}}c}
\toprule
\multirow{2}{*}{\diagbox[width=8em,trim=lr,outerleftsep=0pt, outerrightsep=2pt]{Method}{Target age}} & 0-2    & 3-6    & 7-9    & 15-19  & 30-39  & 50-69  \\ 
\cline{2-7}
\xrowht{5pt}
& \multicolumn{6}{c}{LPIPS$\boldsymbol{\uparrow}$} \\
\midrule
V1 & 0.016 & 0.015 & 0.010 & 0.007 & 0.006 & 0.009 \\
V2 & 0.106 & 0.089 & 0.071 & 0.083 & 0.118 & 0.111 \\
V3 & 0.077 & 0.059 & 0.054 & 0.081 & 0.084 & 0.086 \\
Ours full & \textbf{0.204} & \textbf{0.143} & \textbf{0.132} & \textbf{0.166} & \textbf{0.184} & \textbf{0.182}  \\ 
\bottomrule
\end{tabular}
\end{table}
\textit{Hybrid diversity mechanism.} We use three variants of our method to explore the impact of components in the proposed compounded diversity design. Our $\rm{DLAT}_{img}$ is regarded as the second variant (V2). We obtain the first version (V1) based on V2 by removing the age latent code predictor module and related losses. As a result, the implementation of this version is close to DLFS. Since $\rm{DLAT}_{lmk}$ cannot work independently, we combine it with the V1 model, forming the third variant (V3). From the generations illustrated in Fig. \ref{fig:ablation-global}, we observe that V1 produces deterministic facial rejuvenation. After warping its syntheses with the transformation provided by $\rm{DLAT}_{lmk}$, slight but noticeable variations show up on the facial shape (pay attention to the difference of face counters yielded by V3). On the other hand, directly introducing a diversity mechanism to V1 at the pixel and feature vector levels (i.e., V2) mainly brings uncertainty to the facial texture. Our full model, the integration of $\rm{DLAT}_{img}$ and $\rm{DLAT}_{lmk}$, demonstrates the most comprehensive age transformation diversity, including textural and structural aspects. Relevant objective results in Table \ref{diversity_ablation} further validate the above visual findings.
\par
\begin{figure}[tbp]
\centering
    \makebox[0.10\textwidth]{\scriptsize Input (50-69)}
    \makebox[0.20\textwidth]{\scriptsize Synthesis (30-39)}
    \makebox[0.01\textwidth]{}
    \vspace{2pt}
    \\
    \includegraphics[width=0.10\textwidth]{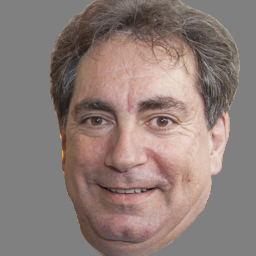}
    \includegraphics[width=0.20\textwidth]{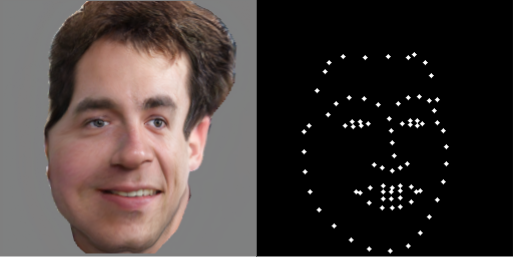}
    \raisebox{0.4\height}{\makebox[0.01\textwidth]{\rotatebox{90}{\makecell{\scriptsize w/o $\mathcal{L}_{pos}$}}}}
    \vspace{3pt}
    \\
    \makebox[0.10\textwidth]{}
    \includegraphics[width=0.20\textwidth]{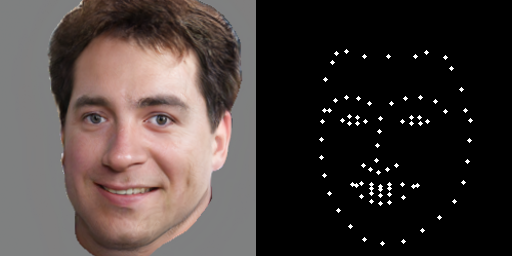}
    \raisebox{0.5\height}{\makebox[0.01\textwidth]{\rotatebox{90}{\makecell{\scriptsize w/ $\mathcal{L}_{pos}$}}}}
    \vspace{3pt}
    \\
\caption{Generations comparison of training our $\rm{DLAT}_{lmk}$ without (w/o) or with (w/) the pose consistency constraint.}
\label{fig:ablation_L_pose}
\end{figure}

\begin{figure}[tbp]
\centering
    \makebox[0.10\textwidth]{\scriptsize Input (30-39)}
    \makebox[0.10\textwidth]{\scriptsize Synthesis (3-6)}
    \makebox[0.10\textwidth]{\scriptsize Synthesis (3-6)}
    \vspace{2pt}
    \\
    \includegraphics[width=0.10\textwidth]{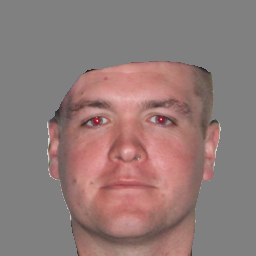}
    \includegraphics[width=0.10\textwidth]{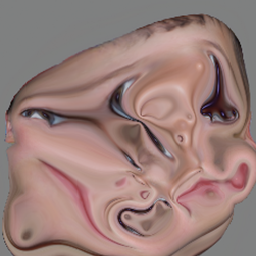}
    \includegraphics[width=0.10\textwidth]{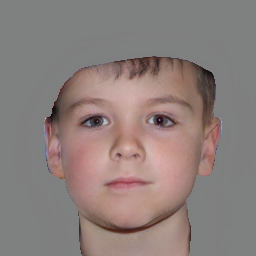}
    \\
    \makebox[0.10\textwidth]{}
    \makebox[0.10\textwidth]{\scriptsize w/o PCA}
    \makebox[0.10\textwidth]{\scriptsize w/ PCA}
    \\
    \caption{Generations comparison of implementing our $\rm{DLAT}_{lmk}$ without (w/o) or with (w/) PCA feature dimension reduction.}
    \label{fig:ablation_pca}
\end{figure}

\textit{Pose and race conformance constraints.} The race consistency constraint is conducted on $\rm{DLAT}_{img}$ by using $\mathcal{L}_{rac}$ loss in its training process. Visual results shown in Fig. \ref{fig:ablation_L_race} demonstrate that without this constraint, the network tends to overly change the skin color to maximize the age transformation diversity, causing unwanted falsification of the subject's ethnic attribute. In contrast, employing this restriction helps the $\rm{DLAT}_{img}$ avoid possible race-changing problems. On the other hand, the pose consistency demand is made to $\rm{DLAT}_{lmk}$ with $\mathcal{L}_{pos}$ loss, aiming at preventing the network from changing the input person's pose to obtain higher variations of landmarks syntheses. Comparison results depicted in Fig. \ref{fig:ablation_L_pose} verify its effectiveness.
\par

\textit{PCA on facial landmarks.} In the realization of $\rm{DLAT}_{lmk}$, we apply PCA to the extracted 2D landmarks and obtain their corresponding 1D feature points. Subsequent feature modulation based on age conditions and diversity mechanism is conducted on these dimensionality-reduced feature points rather than the original landmarks. This operation is because we empirically found it favors the preservation of spatial relationships among individual facial landmarks, while directly processing raw landmarks usually incurs generations with broken spatial structure. A visual evidence is given in Fig. \ref{fig:ablation_pca}.

\par

\section{Discussions}
\begin{table}[tbp]
\caption{Performance of the Face Recognition model Trained on Real images and Addition of Syntheses Generated by LATS and Our Method}
\label{data-extension}
\centering
\begin{tabular}{c@{\hspace{7pt}}c@{\hspace{7pt}}c@{\hspace{7pt}}c@{\hspace{7pt}}c@{\hspace{7pt}}}
\toprule
\multirow{2}{*}{\diagbox[width=8em,trim=lr,outerleftsep=0pt, outerrightsep=2pt]{Training}{Evaluation}} & AgeDB30 & CFP-FP & CFP-FF & CPLFW  \\
\cline{2-5}
\xrowht{5pt}
& \multicolumn{4}{c}{Verification rate$\boldsymbol{\uparrow}$} \\
\midrule
Real & 62.9\% & 6.29\% & 79.9\% & 61.2\% \\
Real + Syn. (LATS) & 63.4\% & 63.5\% & 81.4\% & 64.4\% \\
Real + Syn. (\textbf{ours}) & \textbf{64.8\%} & \textbf{65.0\%} & \textbf{83.0\%} & \textbf{70.0\%} \\
\bottomrule
\end{tabular}
\end{table}


\begin{figure}[tbp]
\centering
    \makebox[0.10\textwidth]{\scriptsize Input (7-9)}
    \hspace{7pt}
    \makebox[0.10\textwidth]{\scriptsize Input (15-19)}
    \vspace{2pt}
    \\
    \includegraphics[width=0.10\textwidth]{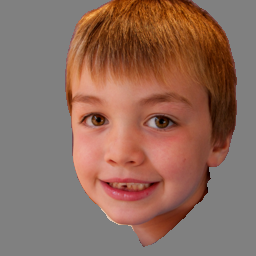}
    \hspace{7pt}
    \includegraphics[width=0.10\textwidth]{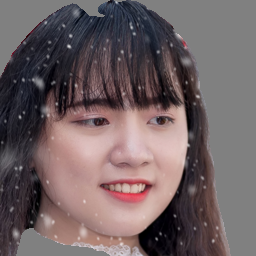}
    \vspace{3pt}
    \\
    \includegraphics[width=0.10\textwidth]{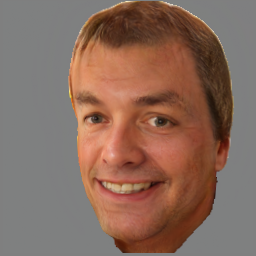}
    \hspace{7pt}
    \includegraphics[width=0.10\textwidth]{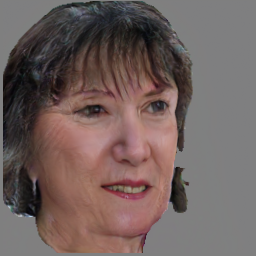}
    \vspace{2pt}
    \\
    \makebox[0.10\textwidth]{\scriptsize Synthesis (30-39)}
    \hspace{7pt}
    \makebox[0.10\textwidth]{\scriptsize Synthesis (50-69)}
    \\
\caption{Representative failure generations yielded by our $\rm{DLAT}^{\boldsymbol{+}}$. These syntheses show grotesqueness in facial structures.}
\label{fig:failure_results}
\end{figure}
\textit{Applications.} It deserves discussion of why the research on how to realize ``diverse and lifespan facial age transformation synthesis" is valuable. Here, we give our arguments from two aspects: the facts and the potential applications. From a factual point of view, human facial aging progress is affected by multiple factors, belonging to internal genetic and external environmental levels. This leads to the uncertainty of aging in reality. Based on this fact, a typical application of our method is to generate multiple possible appearances of the reference face at the specified age condition. Compared to deterministic age transformation, this function should be more helpful in assisting the search for missing children, providing candidates for age-shifting-related visual effects in film production, etc. On the other hand, since the collection of faces is expensive and may cause ethical problems, we are curious about the possibility of expanding real faces with synthesized ones to serve as training data for face recognition algorithms \cite{10097133}. Therefore, we randomly select 10,000 samples from CASIA-WebFace to form the original set. On the other hand, two sets of 24,000 synthetic faces belonging to 1,000 identities are produced by our $\rm{DLAT}^{\boldsymbol{+}}$ and LATS \cite{or2020lifespan}, respectively. Note that we do not include others but choose the LATS as a representative because those alternatives all produce deterministic aging effects while ours generate diverse results. We train the ArcFace model \cite{deng2019arcface} on the original data and its integration with two sets of synthesized data, respectively. Upon convergence, these three versions are evaluated on four public benchmark datasets, including AgeDB \cite{moschoglou2017agedb}, CFP-FP \cite{sengupta2016frontal}, CFP-FF\cite{sengupta2016frontal}, and CPLFW \cite{zheng2018cross}.  Relevant results given in Table \ref{data-extension} show that these synthetic faces with variable age transformations can better boost face recognition performance.

\textit{Limitations.} Although our technique proves its uniqueness and effectiveness in extensive experiments, failures occasionally appear, and some typical samples are given in Fig. \ref{fig:failure_results}. For instance, if the reference portrait is a side face with a relatively large yaw angle relative to the frontal pose, the output generations occasionally seem strange in facial structure. To alleviate this problem, involving prior knowledge of complete 3D heads may be favorable. In addition, replacing the GAN-based generative networks with more powerful Diffusion models may further boost the synthesized image quality. We leave these explorations for our future work.

\section{Conclusion}
This paper presents a new method to synthesize diverse and plausible age transformations on human faces while allowing the target age stages to be chosen throughout the lifetime. Benefiting from the specially designed mechanism, various aging effects simultaneously manifest in the facial texture and shape aspects. Moreover, based on the golden standard summarized from abundant sequential faces of the same persons but at different ages, we also introduce the first metric for assessing the rationality of identity variations between input portraits and corresponding age-regressed or progressed syntheses. We believe all these contributions will pave the way for the further landing of age-transformation techniques.  
\par

\clearpage

\section{Additional Results}
Because of the page's limitation, the main text only exhibits visualized results on a single gender in each experiment. In Fig. \ref{fig:females}, we provide additional example comparisons between our $\rm{DLAT}^{\boldsymbol{+}}$ and other state-of-the-art approaches when the reference face for age transformation belongs to a female. In addition, Fig. \ref{fig:males_diversity} presents diverse lifespan age transformation synthesis on a male face generated by our method.
\begin{figure*}[htbp]
\centering
\renewcommand{\thefigure}{11}
\addtocounter{figure}{+10}
    \makebox[0.10\textwidth]{\scriptsize Input (15-19)}
    \makebox[0.10\textwidth]{\scriptsize 0-2}
    \makebox[0.10\textwidth]{\scriptsize 3-6}
    \makebox[0.10\textwidth]{\scriptsize 7-9}
    \makebox[0.10\textwidth]{\scriptsize 15-19}
    \makebox[0.10\textwidth]{\scriptsize 30-39}
    \makebox[0.10\textwidth]{\scriptsize 50-69}
    \vspace{2pt}
    \\
    \includegraphics[width=0.10\textwidth]{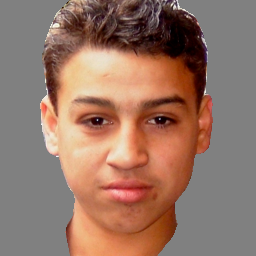}
    \includegraphics[width=0.10\textwidth]{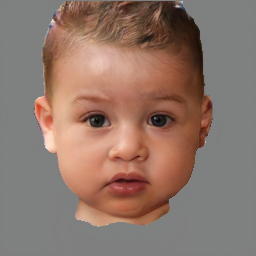}
    \includegraphics[width=0.10\textwidth]{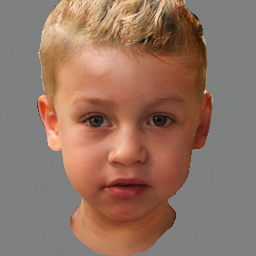}
    \includegraphics[width=0.10\textwidth]{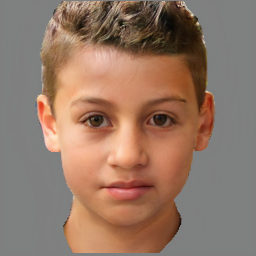}
    \includegraphics[width=0.10\textwidth]{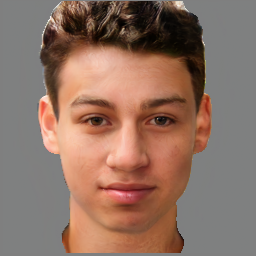}
    \includegraphics[width=0.10\textwidth]{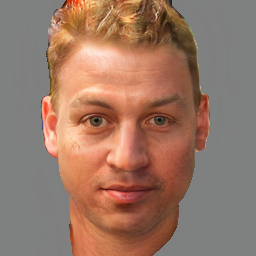}
    \includegraphics[width=0.10\textwidth]{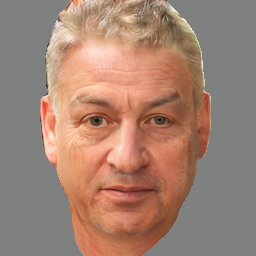}
    \vspace{3pt}
    \\
    \makebox[0.10\textwidth]{}
    \includegraphics[width=0.10\textwidth]{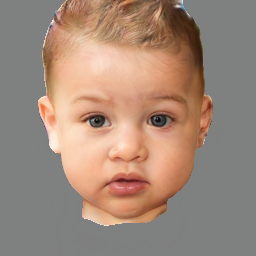}
    \includegraphics[width=0.10\textwidth]{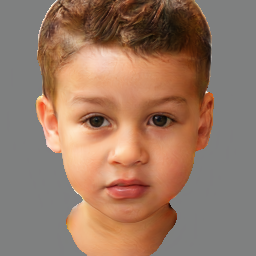}
    \includegraphics[width=0.10\textwidth]{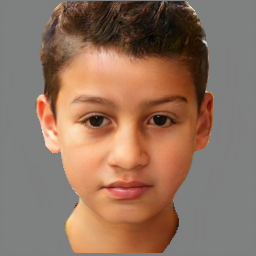}
    \includegraphics[width=0.10\textwidth]{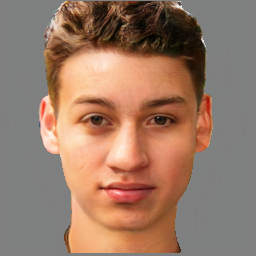}
    \includegraphics[width=0.10\textwidth]{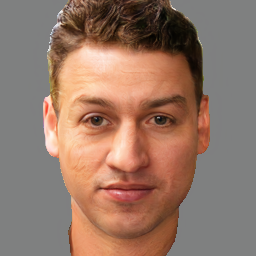}
    \includegraphics[width=0.10\textwidth]{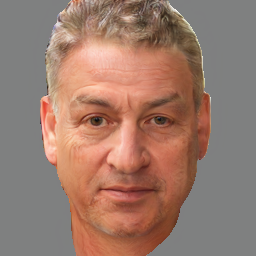}
    \vspace{3pt}
    \\
    \makebox[0.10\textwidth]{}
    \includegraphics[width=0.10\textwidth]{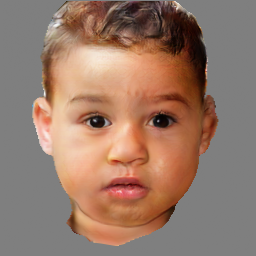}
    \includegraphics[width=0.10\textwidth]{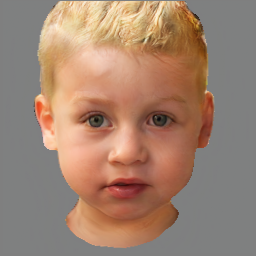}
    \includegraphics[width=0.10\textwidth]{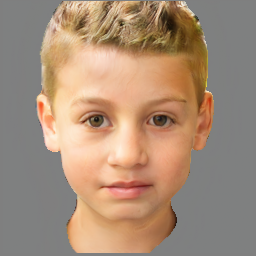}
    \includegraphics[width=0.10\textwidth]{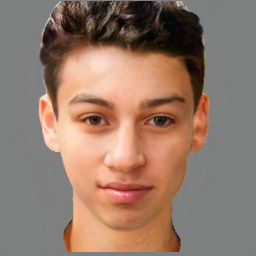}
    \includegraphics[width=0.10\textwidth]{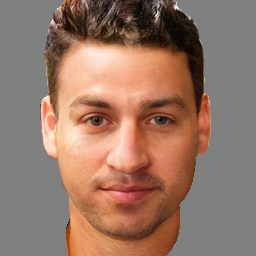}
    \includegraphics[width=0.10\textwidth]{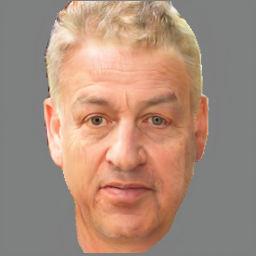}
    \vspace{3pt}
    \\
    \makebox[0.10\textwidth]{}
    \includegraphics[width=0.10\textwidth]{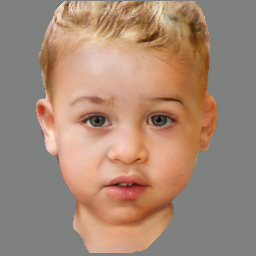}
    \includegraphics[width=0.10\textwidth]{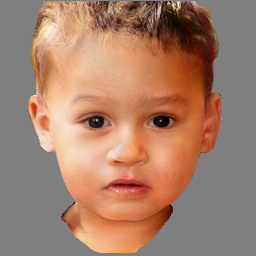}
    \includegraphics[width=0.10\textwidth]{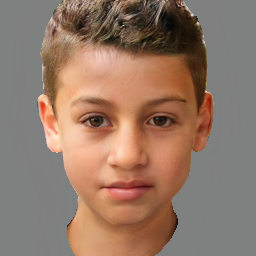}
    \includegraphics[width=0.10\textwidth]{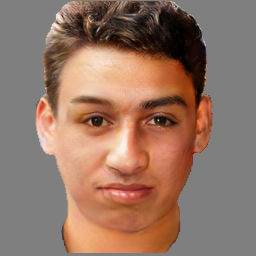}
    \includegraphics[width=0.10\textwidth]{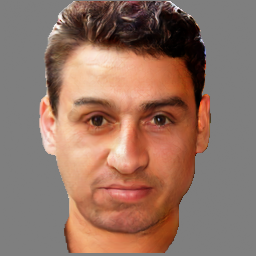}
    \includegraphics[width=0.10\textwidth]{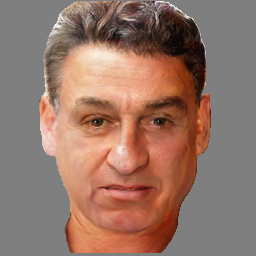}
    \vspace{3pt}
    \\
\caption{Visualization of diverse age transformation results across the life cycle, generated by applying our full method to a male subject.}
\label{fig:males_diversity}
\end{figure*}

\begin{figure*}[htb] \centering
\renewcommand{\thefigure}{12}
    \makebox[0.10\textwidth]{\scriptsize Input (7-9)}
    \makebox[0.10\textwidth]{\scriptsize 0-2}
    \makebox[0.10\textwidth]{\scriptsize 3-6}
    \makebox[0.10\textwidth]{\scriptsize 7-9}
    \makebox[0.10\textwidth]{\scriptsize 15-19}
    \makebox[0.10\textwidth]{\scriptsize 30-39}
    \makebox[0.10\textwidth]{\scriptsize 50-69}
    \makebox[0.01\textwidth]{}
    \vspace{2pt}
    \\
    \includegraphics[width=0.10\textwidth]{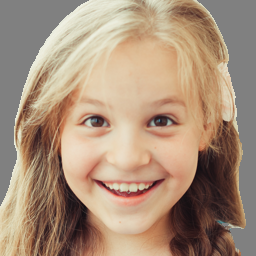}
    \includegraphics[width=0.10\textwidth]{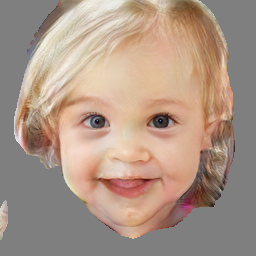}
    \includegraphics[width=0.10\textwidth]{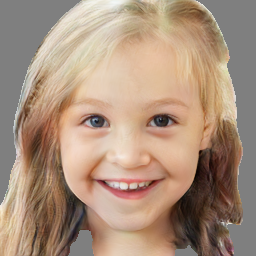}
    \includegraphics[width=0.10\textwidth]{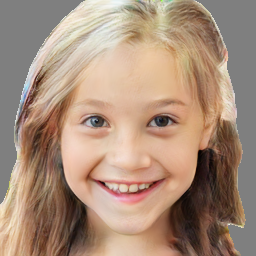}
    \includegraphics[width=0.10\textwidth]{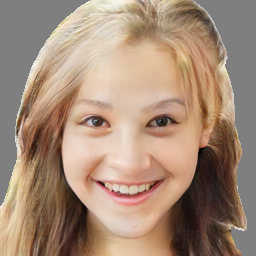}
    \includegraphics[width=0.10\textwidth]{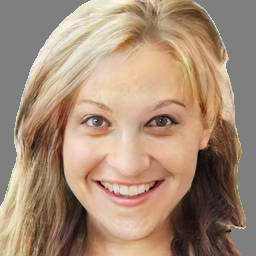}
    \includegraphics[width=0.10\textwidth]{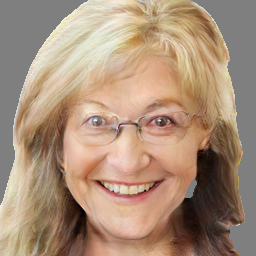}
    \raisebox{1.5\height}{\makebox[0.01\textwidth]{\rotatebox{90}{\makecell{\scriptsize LATS}}}}
    \vspace{3pt}
    \\
    \makebox[0.10\textwidth]{}
    \includegraphics[width=0.10\textwidth]{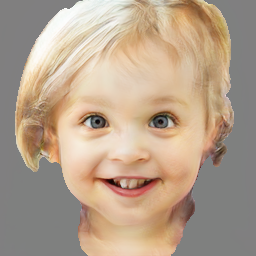}
    \includegraphics[width=0.10\textwidth]{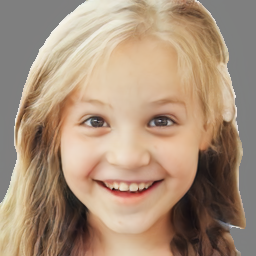}
    \includegraphics[width=0.10\textwidth]{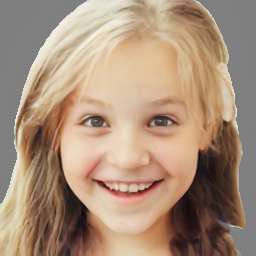}
    \includegraphics[width=0.10\textwidth]{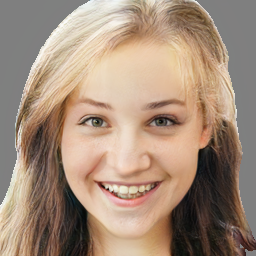}
    \includegraphics[width=0.10\textwidth]{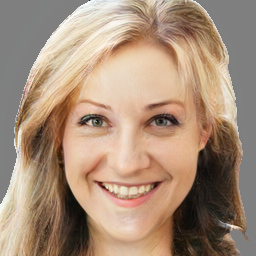}
    \includegraphics[width=0.10\textwidth]{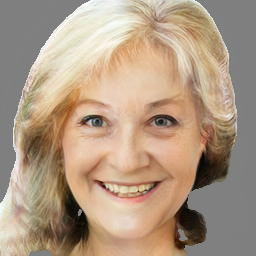}
    \raisebox{1.5\height}{\makebox[0.01\textwidth]{\rotatebox{90}{\makecell{\scriptsize DLFS}}}}
    \vspace{3pt}
    \\
    \makebox[0.10\textwidth]{}
    \includegraphics[width=0.10\textwidth]{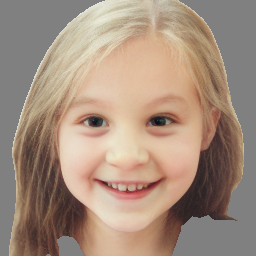}
    \includegraphics[width=0.10\textwidth]{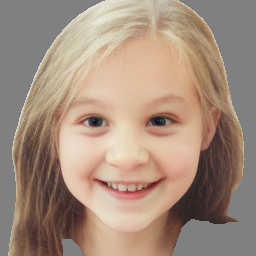}
    \includegraphics[width=0.10\textwidth]{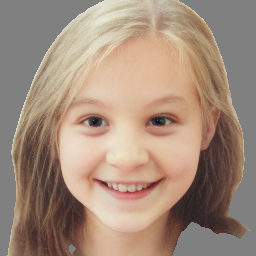}
    \includegraphics[width=0.10\textwidth]{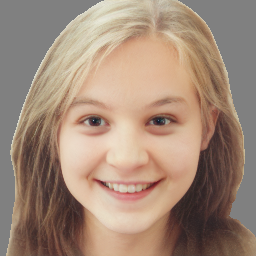}
    \includegraphics[width=0.10\textwidth]{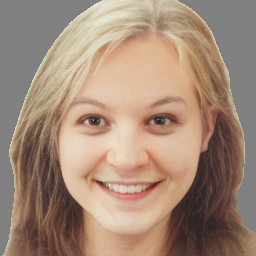}
    \includegraphics[width=0.10\textwidth]{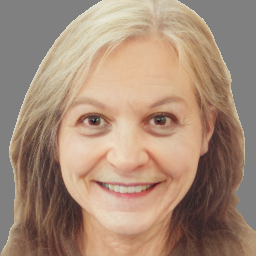}
    \raisebox{1.5\height}{\makebox[0.01\textwidth]{\rotatebox{90}{\makecell{\scriptsize SAM}}}}
    \vspace{3pt}
    \\
    \makebox[0.10\textwidth]{}
    \includegraphics[width=0.10\textwidth]{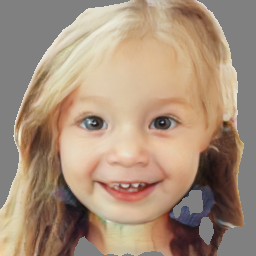}
    \includegraphics[width=0.10\textwidth]{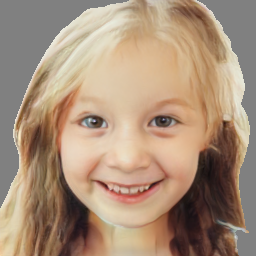}
    \includegraphics[width=0.10\textwidth]{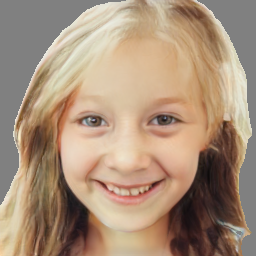}
    \includegraphics[width=0.10\textwidth]{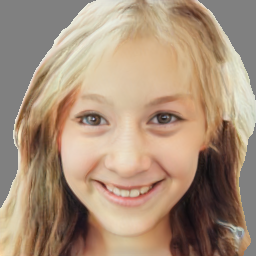}
    \includegraphics[width=0.10\textwidth]{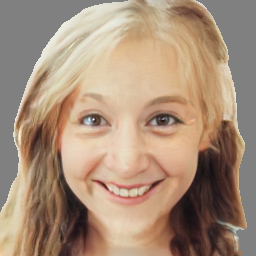}
    \includegraphics[width=0.10\textwidth]{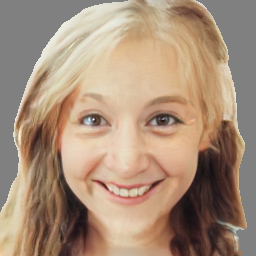}
    \raisebox{1.5\height}{\makebox[0.01\textwidth]{\rotatebox{90}{\makecell{\scriptsize CUSP}}}}
    \vspace{3pt}
    \\
    \makebox[0.10\textwidth]{}
    \includegraphics[width=0.10\textwidth]{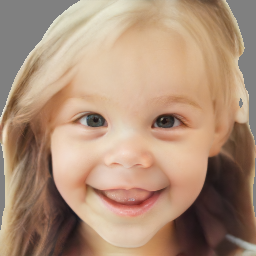}
    \includegraphics[width=0.10\textwidth]{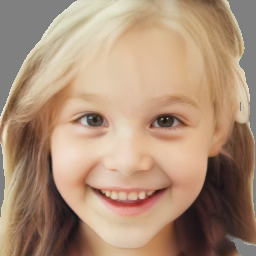}
    \includegraphics[width=0.10\textwidth]{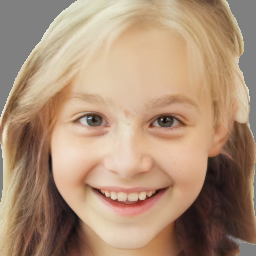}
    \includegraphics[width=0.10\textwidth]{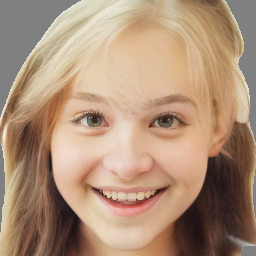}
    \includegraphics[width=0.10\textwidth]{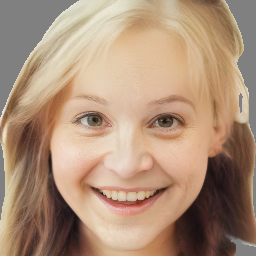}
    \includegraphics[width=0.10\textwidth]{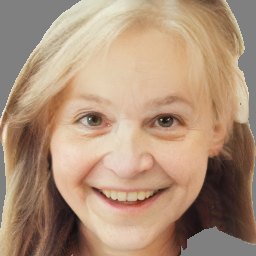}
    \raisebox{0.3\height}{\makebox[0.01\textwidth]{\rotatebox{90}{\makecell{\scriptsize AgeTransGAN}}}}
    \vspace{3pt}
    \\
    \makebox[0.10\textwidth]{}
    \includegraphics[width=0.10\textwidth]{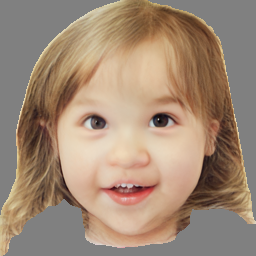}
    \includegraphics[width=0.10\textwidth]{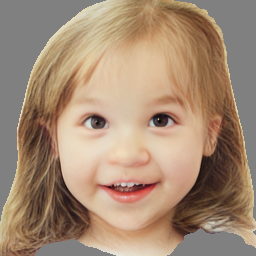}
    \includegraphics[width=0.10\textwidth]{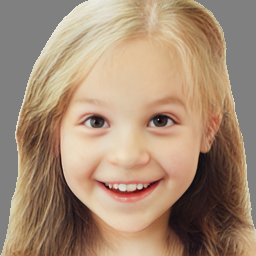}
    \includegraphics[width=0.10\textwidth]{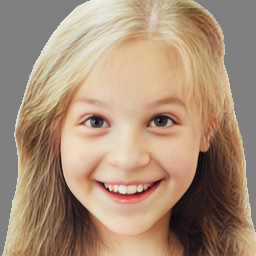}
    \includegraphics[width=0.10\textwidth]{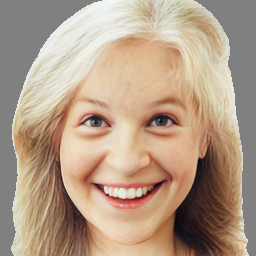}
    \includegraphics[width=0.10\textwidth]{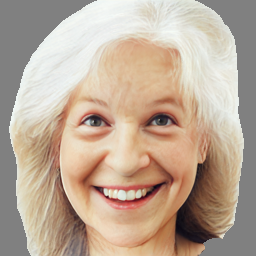}
    \raisebox{0.3\height}{\makebox[0.01\textwidth]{\rotatebox{90}{\makecell{\scriptsize InterFaceGAN}}}}
    \vspace{3pt}
    \\
    \makebox[0.10\textwidth]{}
    \includegraphics[width=0.10\textwidth]{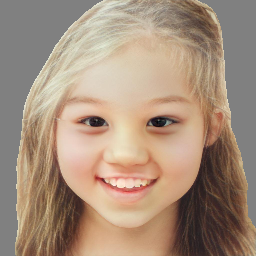}
    \includegraphics[width=0.10\textwidth]{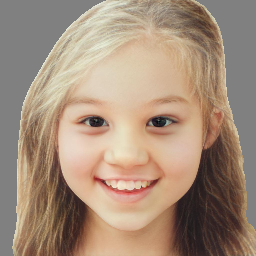}
    \includegraphics[width=0.10\textwidth]{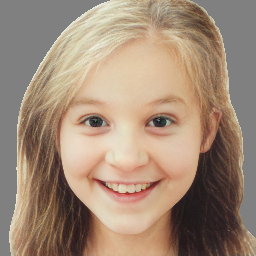}
    \includegraphics[width=0.10\textwidth]{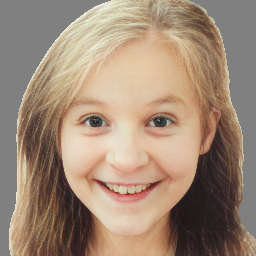}
    \includegraphics[width=0.10\textwidth]{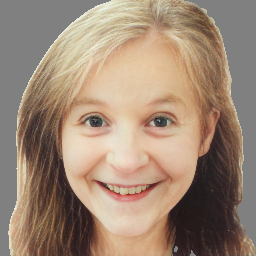}
    \includegraphics[width=0.10\textwidth]{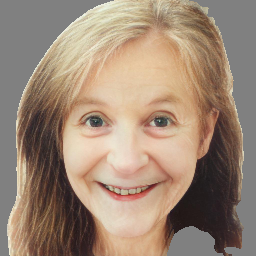}
    \raisebox{0.1\height}{\makebox[0.01\textwidth]{\rotatebox{90}{\makecell{\scriptsize Latent-Transformer}}}}
    \vspace{3pt}
    \\
    \makebox[0.10\textwidth]{}
    \includegraphics[width=0.10\textwidth]{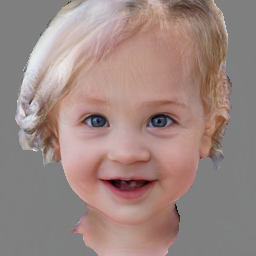}
    \includegraphics[width=0.10\textwidth]{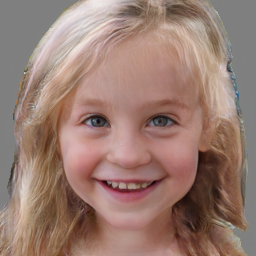}
    \includegraphics[width=0.10\textwidth]{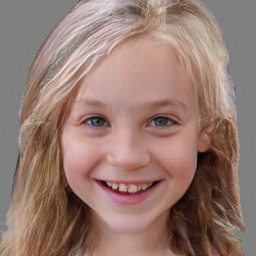}
    \includegraphics[width=0.10\textwidth]{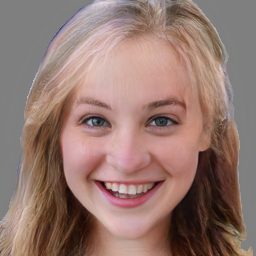}
    \includegraphics[width=0.10\textwidth]{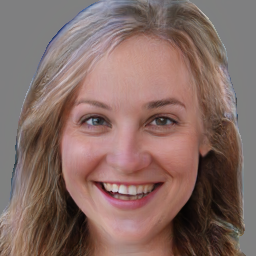}
    \includegraphics[width=0.10\textwidth]{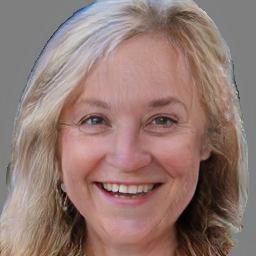}
    \raisebox{0.2\height}{\makebox[0.01\textwidth]{\rotatebox{90}{\makecell{\scriptsize $\rm{DLAT}^{\boldsymbol{+}}$ (ours)}}}}
    \vspace{3pt}
    \\
    \caption{Visual comparison among lifespan age transformation results of the given female subject, which are synthesized by LATS, DLFS, SAM, CUSP, AgeTransGAN, InterFaceGAN, Latent-Transformer and our $\rm{DLAT}^{\boldsymbol{+}}$.} 
    \label{fig:females}
\end{figure*}

\section{Details of Fig. 3}
Fig. 3 of the main text summarizes the statistical results of average scores about identity similarity among faces from the same subject. However, these figures only present trends without exact numerical values, which Table \ref{similarity_scores} exposes here. Note that awareness of these values is necessary to calculate the metric of \textbf{I}dentity \textbf{D}eviation under \textbf{A}ge \textbf{G}aps (IDAG).

\begin{table*}[htbp]
\renewcommand{\thetable}{\uppercase\expandafter{\romannumeral7}} 
\caption{Statistics of Average Similarity Scores among Real Faces of the Same Subjects}
\centering
\footnotesize
\begin{tabular}{c@{\hspace{7pt}}c@{\hspace{7pt}}c@{\hspace{7pt}}c@{\hspace{7pt}}c@{\hspace{7pt}}c@{\hspace{7pt}}c@{\hspace{7pt}}c@{\hspace{7pt}}c@{\hspace{7pt}}c@{\hspace{7pt}}c}
\toprule
\multirow{2}{*}{\diagbox[width=8em,trim=lr,outerleftsep=0pt, outerrightsep=2pt]{Target age}{Source age}} & 0-2 & 3-6 & 7-9 & 10-15 & 15-19  & 20-29  & 30-39  & 40-49  & 50-69  & 70+  \\ 
\cline{2-11}
\xrowht{5pt}
& \multicolumn{10}{c}{Similarity score ($\%$)} \\
\midrule  
0-2  & 100.00  & 80.24  & 71.55  & 63.75  & 52.81   & 46.45  & 43.44  & 41.54  & 40.68  & 39.69 \\
3-6 & 80.24  & 100.00  & 82.83  & 77.64 & 67.68  & 60.77 &55.68 &53.01 &51.85 &50.23  \\
7-9 & 71.55  &82.83  &100.00  &82.64 &74.80  &68.43 &62.73 &59.69 &55.90 &53.37  \\
10-14  &63.75 &77.64  &82.64  &100.00 &80.86  &75.00 &69.87 &66.75 &62.61 &64.37  \\
15-19 &52.81 &67.68 &74.80 &80.86 &100.00 &83.32 &78.97 &75.68 &71.57 &68.10 \\
20-29 &46.45 &60.77 &68.43  &74.99  &83.32  &100.00  &85.39  &81.90  &77.38  &72.31  \\ 
30-39  &43.44 &55.68 &62.73  &69.87  &78.97  &85.39  &100.00  &86.76  &82.22  &75.40  \\
40-49  &41.53 &53.01 &59.69  &66.75  &75.68  &81.90  &86.76  &100.00  &85.69  &78.62 \\ 
50-69  &40.68 &51.85 &55.90  &62.61  &71.57  &77.38  &82.22  &85.69  &100.00  &84.17  \\ 
70+ &39.69 &50.23 &53.37  &64.37  &68.10  &72.31  &75.40  &78.62  &84.17  &100.00  \\ 
\bottomrule
\end{tabular}
\label{similarity_scores}
\end{table*}

\section{Network Architectures}
Our full framework consists of two diverse and lifelong age transformation networks for facial appearance-level and geometry-level synthesis, namely $\rm{DLAT}_{img}$ and $\rm{DLAT}_{lmk}$, respectively. Each network is composed of four modules: a diverse age mapper, an age latent code predictor, a generator, and a discriminator. Next, we elaborate on the architectural details of each individual.

\textit{Diverse age mapper.} 
The architecture of the diverse age mapping module $M_{img}$ for the image network is shown in Table \ref{tab:diverse age mapping network-img}, which is a special multi-layer perceptron. It takes a noise vector randomly sampled from the Gaussian distribution as the input. This signal is first processed by a shared backbone and then fed into $K$ parallel branches, where $K$ is equivalent to the number of age groups. The mapper eventually outputs a latent age code for each age stage. $M_{lmk}$ (Table \ref{tab:diverse age mapping network-lmk}) for the landmark network has a similar structure but with slight differences in the number of hidden layers and the dimension of output age codes (256 vs. 64). In these two mappers, each linear layer except the last one is followed by a ReLU activation.
\begin{table}[tbp]
\renewcommand{\thetable}{\uppercase\expandafter{\romannumeral8}} 
    \centering
    \caption{Architecture of Diverse Age Mapper $M_{img}$}
    \label{tab:diverse age mapping network-img}
        \footnotesize
        \begin{tabular}[t]{*{6}{c}}
        \toprule
        Type & Layer & Activation & Output Shape  \\
        \midrule
        Shared &Noise  &-  & $16$ \\
        \midrule
        Shared &Linear      & ReLU    & $512$      \\
        Shared &Linear      & ReLU    & $512$      \\
        Shared &Linear      & ReLU    & $512$      \\
        Shared &Linear      & ReLU    & $512$      \\
        Shared &Linear      & ReLU    & $512$      \\
        Shared &Linear      & ReLU    & $512$      \\
           
        \midrule
        Unshared  &Linear  &ReLU   &$512$ \\
        Unshared  &Linear  &-   &$256$ \\
       
        \bottomrule
  
        \end{tabular}
\end{table}

\begin{table}[tbp]
\renewcommand{\thetable}{\uppercase\expandafter{\romannumeral9}} 
    \centering
    \caption{Architecture of Diverse Age Mapper $M_{lmk}$.}
    \label{tab:diverse age mapping network-lmk}
        \footnotesize
        \begin{tabular}[t]{*{6}{c}}
        \toprule
        Type & Layer & Activation & Output Shape  \\
        \midrule
        Shared &Noise  &-  &$16$  \\
        \midrule
        Shared &Linear      & ReLU    & 512      \\
        Shared &Linear      & ReLU     & 512      \\
        Shared &Linear      & ReLU     & 512       \\
        Shared &Linear      & ReLU     & 512      \\       
        \midrule
        Unshared  &Linear  &ReLU   &512 \\
        Unshared  &Linear  &ReLU   &512 \\
        Unshared  &Linear  &ReLU   &512 \\
        Unshared  &Linear  &-   &64 \\
        \bottomrule
  
        \end{tabular}
\end{table}

\textit{Age latent code predictor.} Table \ref{tab:age latent code predictor-img} presents the architecture of our age latent code predictor $P_{img}$ in the image-based transformation network, which also has a ``shared-unshared" structure design. To be concrete, the shared part consists of two convolutional layers and six residual blocks, each employing a leaky ReLU as the activation function. Every unshared branch is composed of a single linear layer, and $K$ of them is attached to the shared backbone. On the other hand, details of the $P_{lmk}$ are given in Table \ref{tab:Landmarks Age Latent code predictor}. Note that it takes the vector-formed representation of facial landmarks as the input, which is obtained by applying principal component analysis (PCA) to them. 

\begin{table}[tbp]
\renewcommand{\thetable}{\uppercase\expandafter{\romannumeral10}} 
    \centering
    \caption{Architecture of Age Latent Code Predictor $P_{img}$.}
    \label{tab:age latent code predictor-img}
        \footnotesize
        \begin{tabular}[t]{*{6}{c}}
        \toprule
        Type & Layer & Activation & Output Shape  \\
        \midrule
        Shared  &Image & - &$256\times256\times3$  \\
        \midrule
        Shared  &Conv$1\times1$ & LReLU      &$256\times256\times64$      \\
        Shared  &ResBlk  & LReLU   & $128\times128\times128$      \\
        Shared  &ResBlk   & LReLU  & $64\times64\times256$       \\
        Shared  &ResBlk  & LReLU   & $32\times32\times512$      \\      
        Shared  &ResBlk  & LReLU   & $16\times16\times512$      \\  
        Shared  &ResBlk  & LReLU   & $8\times8\times512$      \\  
        Shared  &ResBlk  & LReLU   & $4\times4\times512$      \\  
        Shared  &Conv$4\times4$  & LReLU      &$1\times1\times512$      \\
        \midrule
        Unshared  &Reshape  & -  & $512$      \\  
        Unshared   &Linear  & -  &$256$ \\
       
        \bottomrule
  
        \end{tabular}
\end{table}

\begin{table}[tbp]
\renewcommand{\thetable}{\uppercase\expandafter{\romannumeral11}} 
    \centering
    \caption{Architecture of Age Latent Code Predictor $P_{lmk}$}
    \label{tab:Landmarks Age Latent code predictor}
        \footnotesize
        \begin{tabular}[t]{*{6}{c}}
        \toprule
        Type & Layer & Activation & Output Shape  \\
        \midrule
        Shared &PCA Landmarks  &-  &32  \\
        \midrule
        Shared &Linear      & LReLU     & 512      \\
        Shared &Linear      & LReLU     & 512      \\
        Shared &Linear      & LReLU     & 512       \\
        Shared &Linear      & LReLU     & 512      \\ 
        Shared &Linear      & LReLU     & 512      \\
        Shared &Linear      & LReLU     & 512     \\
        Shared &Linear      & LReLU     & 512     \\
        Shared &Linear      & LReLU     & 512     \\
        Shared &Linear      & LReLU     & 512     \\
        Shared &Linear      & LReLU     & 512     \\
        \midrule
        Unshared  &Linear  &-   &64 \\
        \bottomrule
  
        \end{tabular}
\end{table}

\textit{Generator.} Our image generator can be regarded as integrating an encoder and a decoder, the architecture of which is shown in Table \ref{tab:img generator}. Specifically, the encoder part contains three convolutional layers and four residual blocks, each followed by a ReLU and pixel-normalization \cite{karras2018progressive}. We feed the reference image into them and obtain a corresponding feature map. Regarding the decoder part, we use six styled convolution blocks \cite{karras2019style} to modulate the feature map with age latent codes and bilinear upsampling in the last two blocks to restore the original resolution. Here, the activation function used changes to the leaky ReLU. After that, a convolutional layer is leveraged to produce the final age-transformed facial image. Details of the landmark generator $G_{lmk}$ are summarized in Table \ref{tab:Landmarks G}. We employ eight FiLM layers \cite{perez2018film} to modulate PCA-processed landmarks with age latent codes.
\begin{table}[tbp]
\renewcommand{\thetable}{\uppercase\expandafter{\romannumeral12}} 
    \centering
    \caption{Architecture of Image Generator $G_{img}$}
    \label{tab:img generator}
        \footnotesize
        \begin{tabular}[t]{*{6}{c}}
        \toprule
        Layer  & Activation & Norm & Output Shape  \\
        \midrule
        Image   &-   &-  &$256\times256\times3$ \\
        \hline 
        Conv$7\times7$    &ReLU   &Pixel    & $256\times256\times64$      \\
        Conv$3\times3$    &ReLU   &Pixel    & $128\times128\times128$      \\
        Conv$3\times3$    &ReLU   &Pixel    & $64\times64\times256$      \\
        ResnetBlock    &ReLU   &Pixel    & $64\times64\times256$      \\
        ResnetBlock    &ReLU   &Pixel    & $64\times64\times256$      \\
        ResnetBlock    &ReLU   &Pixel    & $64\times64\times256$      \\
        ResnetBlock    &ReLU   &Pixel    & $64\times64\times256$      \\
        ResnetBlock    &ReLU   &Pixel    & $64\times64\times256$      \\
        StyledConv     &LReLU  &Pixel   & $64\times64\times256$      \\
        StyledConv     &LReLU  &Pixel   & $64\times64\times256$      \\
        StyledConv     &LReLU  &Pixel   & $64\times64\times256$      \\
        StyledConv     &LReLU  &Pixel   & $64\times64\times256$      \\
        StyledConv     &LReLU  &Pixel   & $64\times64\times128$      \\
        Upsample        &-           &-           &$128\times128\times128$    \\
        StyledConv     &LReLU  &Pixel   & $128\times128\times64$      \\
        Upsample        &-           &-           &$256\times256\times64$    \\
        Conv$1\times1$           &Tanh        &-           &$256\times256\times3$     \\
        \bottomrule
  
        \end{tabular}
\end{table}

\begin{table}[tbp]
\renewcommand{\thetable}{\uppercase\expandafter{\romannumeral13}} 
    \centering
    \caption{Architecture of Landmarks Generator $G_{lmk}$}
    \label{tab:Landmarks G}
        \footnotesize
        \begin{tabular}[t]{*{6}{c}}
        \toprule
        Layer & Activation & Output Shape  \\
        \midrule
        PCA Landmarks  &-  &32  \\
        \midrule
        FiLMLayer   & LReLU     & 256      \\
        FiLMLayer   & LReLU     & 256     \\
        FiLMLayer   & LReLU     & 256      \\
        FiLMLayer   & LReLU     & 256     \\ 
        FiLMLayer   & LReLU     & 256     \\
        FiLMLayer   & LReLU     & 256    \\
        FiLMLayer   & LReLU     & 256    \\
        FiLMLayer   & LReLU     & 256    \\
        Linear      & -             & 32    \\
        \bottomrule
  
        \end{tabular}
\end{table}

\textit{Discriminator.} Since the number of age groups $K$ is more than one, two discriminators in our framework both serve for multi-task discrimination. The architecture of the image discriminator $D_{img}$ is given in Table \ref{tab:img Discrimninator}, which mimics the design of StyleGAN. We regularly use average pooling to downsample intermediate feature maps. In addition, mini-batch standard deviation \cite{karras2018progressive} is added before the last convolutional block. The activation used throughout the discriminator is the leaky ReLU. Details of the landmark discriminator $D_{lmk}$ are shown in Table \ref{tab:landmark discriminator}. Here, we again adopt the ``shared-unshared" network structure, where the quantity of unshared branches is equals $K$.

\begin{table}[tbp]
\renewcommand{\thetable}{\uppercase\expandafter{\romannumeral14}} 
    \centering
    \caption{Architecture of Image Discriminator $D_{img}$}
    \label{tab:img Discrimninator}
        \footnotesize
        \begin{tabular}[t]{*{6}{c}}
        \toprule
        Layer  & Activation & Output Shape  \\
        \midrule
        Image  &-  &$256\times256\times3$ \\
        \midrule
        Conv$1\times1$       &LReLU    & $256\times256\times64$      \\
        Conv$3\times3$       &LReLU    & $256\times256\times64$      \\
        Conv$3\times3$       &LReLU    & $256\times256\times128$      \\
        Downsample  &-         & $128\times128\times128$  \\
        Conv$3\times3$       &LReLU    & $128\times128\times128$      \\
        Conv$3\times3$       &LReLU    & $128\times128\times256$      \\
        Downsample  &-         & $64\times64\times256$  \\
        Conv$3\times3$       &LReLU    & $64\times64\times256$      \\
        Conv$3\times3$       &LReLU    & $64\times64\times512$      \\
        Downsample  &-         & $32\times32\times512$  \\
        Conv$3\times3$       &LReLU    & $32\times32\times512$      \\
        Conv$3\times3$       &LReLU    & $32\times32\times512$      \\
        Downsample  &-         & $16\times16\times512$  \\
        Conv$3\times3$       &LReLU    & $16\times16\times512$      \\
        Conv$3\times3$       &LReLU    & $16\times16\times512$      \\
        Downsample  &-         & $8\times8\times512$  \\
        Conv$3\times3$       &LReLU    & $8\times8\times512$      \\
        Conv$3\times3$       &LReLU    & $8\times8\times512$      \\
        Downsample  &-         & $4\times4\times512$  \\
        Minibatch Stdev  &-    &$4\times4\times512$ \\
        Conv$3\times3$       &LReLU    & $4\times4\times512$      \\
        Conv$4\times4$       &LReLU    &$1\times1\times K$      \\
    
        \bottomrule
  
        \end{tabular}
\end{table}

\begin{table}[tbp]
\renewcommand{\thetable}{\uppercase\expandafter{\romannumeral15}} 
    \centering
    \caption{Architecture of Landmarks Discriminator $D_{lmk}$.}
    \label{tab:landmark discriminator}
        \footnotesize
        \begin{tabular}[t]{*{6}{c}}
        \toprule
        Type & Layer & Activation & Output Shape  \\
        \midrule
        Shared &PCA Landmarks  &-  &32  \\
        \midrule
        Shared &Linear      & LReLU    & 256      \\
        Shared &Linear      & LReLU     & 256      \\
        Shared &Linear      & LReLU     & 256       \\
        Shared &Linear      & LReLU     & 256      \\       
        \midrule
        Unshared  &Linear  &LReLU   &256 \\
        Unshared  &Linear  &LReLU   &256 \\
        Unshared  &Linear  &LReLU   &128 \\
        Unshared  &Linear  &Sigmoid   &1 \\
        \bottomrule
  
        \end{tabular}
\end{table}


\vfill

\clearpage

\bibliographystyle{IEEEtran}
\bibliography{IEEEabrv,references}

\end{document}